\documentclass[letterpaper]{article} 
\usepackage{aaai23}  
\usepackage{times}  
\usepackage{helvet}  
\usepackage{courier}  
\usepackage[hyphens]{url}  
\usepackage{graphicx} 
\usepackage{natbib}  
\usepackage{caption} 
\nocopyright
\usepackage{amssymb}
\usepackage{amsthm}
\usepackage{mathtools}
\usepackage{qtree}
\usepackage[ruled,vlined]{algorithm2e}

\newtheorem{definition}{Definition}
\newtheorem{theorem}{Theorem}

\newtheorem{lemma}[theorem]{Lemma}

\newcommand{\citepw}[1]{\citeauthor{#1}~(\citeyear{#1})}

\usepackage{color}

\title{A Formal Metareasoning Model of Concurrent Planning and Execution
}

\author{
Amihay Elboher\textsuperscript{\rm 1},
Ava Bensoussan\textsuperscript{\rm 1},
Erez Karpas\textsuperscript{\rm 2},\\
Wheeler Ruml\textsuperscript{\rm 3},
Shahaf S. Shperberg\textsuperscript{\rm 1},
Solomon E. Shimony\textsuperscript{\rm 1}
}
\affiliations {
\textsuperscript{\rm 1}Ben-Gurion University, Israel
\\\textsuperscript{\rm 2}Technion, Israel\\
\textsuperscript{\rm 3}University of New Hampshire, USA\\
\{amihaye,avabe,shperbsh\}@post.bgu.ac.il, karpase@technion.ac.il, 
shimony@cs.bgu.ac.il, ruml@cs.unh.edu
}

\usepackage{mdframed} 
\usepackage{xparse} 
\usepackage[inline]{enumitem}

\newmdenv[ 
  topline=false,
  bottomline=false,
  rightline=false,
    innertopmargin = 0pt,
    innerleftmargin = 3pt,
    innerrightmargin = 0pt,
    innerbottommargin = 0pt
]{leftrule}

\newcounter{examplecounter}

\NewDocumentEnvironment{example}{O{
\refstepcounter{examplecounter}
\textbf{Example \theexamplecounter:}}}
{\begin{leftrule}\noindent{#1}}
{\end{leftrule}}

\begin{document}
	
\maketitle

\begin{abstract}
Agents that plan and act in the real world must deal with the fact that time passes as they are planning. When timing is tight, there may be insufficient time to complete the search for a plan before it is time to act.  By commencing execution before search concludes, one gains time to search by making planning and execution concurrent. However, this incurs the risk of making incorrect action choices, especially if actions are irreversible. This tradeoff between opportunity and risk is the problem addressed in this paper.
Our main contribution is to formally define this setting as an abstract metareasoning problem.
We find that the abstract problem is intractable.  However, we identify special cases that are solvable in polynomial time, develop greedy solution algorithms, and, through tests on instances derived from search problems, find several methods that achieve promising practical performance.  This work lays the foundation for a principled time-aware executive that concurrently plans and executes.
\end{abstract}

\section{Introduction}

In the real world, time passes as agents plan. For example, an agent that needs to get to the airport may have two options: take a taxi or ride a commuter train. Each of these options can be thought of as a {\em partial plan} to be elaborated into a complete plan before execution can start.
Clearly, the agent's planner should only elaborate the partial plan that involves the train if that can be done before the train leaves.  Note, however, that in general this may require delicate scheduling of search effort across multiple competing partial plans.
Elaborating the example, suppose the planner has two partial plans that are each estimated to require five minutes of computation to elaborate into complete plans. If only six minutes remain until they both expire, then we would want the planner to allocate almost all of its remaining planning effort to one of them, rather than to fail on both.  Issues like these have been the focus of previous work on situated temporal planning \cite{DBLP:conf/aips/CashmoreCCKMR18,aaai19paper}.
	
In this paper, we consider the design of a bolder agent that can begin execution of actions before a complete plan to a goal has been found.  Consider a further extension of the example in which the estimated time to complete each plan is seven minutes. The only apparent way to achieve the agent's goal thus involves starting to act before planning is complete. However, when an action is executed, plans that are not consistent with this action may become invalid, thus incurring the risk of never reaching the goal. This risk is most significant when actions are irreversible. However, even when an action can be reversed, the time spent executing a reversible action (and potentially its inverse action) might still cause the invalidation of plans.
For example, taking the commuter train invalidates the partial plan of taking a taxi, as there will not be enough time to take the train in the opposite direction to still catch the
taxi and reach the airport before the deadline. Thus, if the planner fails to elaborate the partial plan of riding the train into a complete plan that reaches the airport on time, the agent will miss its flight.
This paper proposes a disciplined method for making execution decisions while handling such tradeoffs when allocating search effort in situated planning.
	
The idea of starting to perform actions in the
real world (also known as base-level actions)
before completing the search goes back at least
as far as Korf's (\citeyear{korf:rth}) real-time A*.
The difference from the real-time search setting is that our scenario is more flexible: the agent does not
have a predefined time at which the next action  must be executed.
Rather, it can choose when base-level actions should be executed in order to maximize the probability of
successful and timely execution. Note that we assume that the world is deterministic.
The only uncertainty we model concerns how long planning will take and the time it will take the 
as-yet-unknown resulting plan to reach a goal state, i.e., we only consider uncertainty at the meta-level.  Our setting is also different from the interleaving of planning and execution in order to account for stochastic actions or partial observability, which has been a part of practical applications of planning since the early days of Shakey the robot \cite{DBLP:journals/ai/FikesHN72} and later (e.g., \citet{
ambros1988integrating,haigh1998interleaving,lemai2004interleaving,nourbakhsh2012interleaving}).

Our main contribution is defining
the above issues as a formal problem of decision-making under uncertainty, in the sense defined by \citepw{DBLP:journals/ai/RussellW91}.
Attempting this formalization for a full realistic search algorithm appears daunting,
even under our assumption of a deterministic world.
We thus begin from the formal (AE)$^2$ model (for Allocating Effort when Actions Expire) of \citepw{aaai19paper}, which formalizes situated planning as an abstract metareasoning problem of allocating processing time among $n$ search processes (e.g., subtrees).  The objective is
to find a policy that maximizes the
probability of finding a solution plan that is still feasible to execute when it is found.  We extend the model to allow execution of actions in the real world
{\em in parallel} with the search processes.  We call this model Concurrent Planning
and Execution (CoPE for short).

CoPE is a generalization of (AE)$^2$, so
finding an optimal CoPE policy is also intractable,
even under the assumption of known deadlines and remainders.
Still, we cast CoPE as an MDP, so that we can define and analyze optimal policies,
and even solve CoPE optimally for very small instances using standard MDP techniques like value iteration.  We then describe several efficient suboptimal ways of solving CoPE and evaluate them empirically. 
We find that our 
algorithms span a useful range of speed / effectiveness trade-offs.

This paper examines the static version of the metareasoning problem, i.e. solving the CoPE
instance as observed at a snapshot
of the planning process.
Using our results in a temporal planner would likely involve gathering the requisite statistics and solving CoPE repeatedly, possibly after each node expansion. These integration issues are important future work that is beyond the scope of the current paper.

\section{Background
} \label{sec:background}

In situated temporal planning~\cite{DBLP:conf/aips/CashmoreCCKMR18},
each possible action $a$ (i.e., an action in the search tree that the agent can choose whether to execute or not) has a latest start time $t_a$ and a plan must be fully generated before its first
action can begin executing.
This induces a planning deadline which might be unknown, since the
actions in the plan are not known until the search terminates.
For a partial plan available at a search node $i$ in the planner,
the unknown deadline by which any potential plan expanded from node $i$ must start executing can be modeled by a random
variable.  Thus, the planner faces the metareasoning problem of deciding which nodes
on the open list to expand in order to maximize the chance
of finding a plan before its deadline.

\citet{aaai19paper}\ propose the (AE)$^2$ model, which abstracts away from the search details and merely posits $n$ independent `processes.'
Each process is attempting to solve the same problem under time constraints. In the context of situated temporal planning using heuristic search, each process may represent a promising partial plan for the goal, implemented as a node on the open list, where the allocation of CPU time to that process is equivalent to the exploration of the subtree under the corresponding node.
But the abstract problem may also be applicable  to
other settings, such as algorithm portfolios or scheduling candidates for job interviews.
The metareasoning problem is to determine how to schedule the $n$ processes.

When process $i$ terminates, it delivers a solution with probability $P_i$ or,  otherwise, indicates
its failure to find one.
As mentioned above, each process has an uncertain deadline, defined over
absolute wall-clock time, by which its computation must be
completed in order for any solution it finds to be valid. 
For process $i$, $D_i(t)$ denotes the CDF over wall 
clock times of the
random variable denoting the latest start time (deadline).
This value is only
discovered with certainty when the process completes.
This models the fact that a plan's dependence on an external timed event, such as a train departure, might not become clear until the final action in a plan is added. If a process terminates with a solution before its deadline, it is called {\em timely}.  Given $D_i(t)$, one can assume w.l.o.g. that $P_i$ is 1, otherwise adjust $D_i(t)$ to make the probability of a deadline that is
in the past (thus forcing the plan to fail) equal to $1-P_i$.

The processes have known search time distributions, i.e. performance
profiles \cite{Zilberstein} described by CDFs $M_i(t)$, 
the probability that process $i$ 
needs total computation time $t$
or less to terminate.
As is typical in metareasoning,
(AE)$^2$ assumes that all the random
variables are independent.
Given the $D_i(t)$ and $M_i(t)$ 
distributions, 
the objective of (AE)$^2$ is to schedule processing time between the $n$ processes
such that the probability of at least one process finding a timely solution is maximized. 

A simplified discrete-time version of the problem, called S(AE)$^2$, was cast as a Markov decision process.
The MDP's actions are to assign (schedule) the next time unit to process $i$, denoted by $c_i$ with $i\in [1,n]$.
Computational action $c_i$ is allowed only if process $i$ has not already failed.
A process is considered to have failed
if it has terminated and discovered that its
deadline has already passed, or if
the current time is later than the last possible deadline for the process. Transitions are determined by the probabilistic performance profile $M_i$ and the deadline distributions (see \citet{aaai19paper} for details).
A process terminating with a timely solution
results in success and a reward of 1.

\begin{example} 
We need
to get to terminal C at
the airport 30 minutes from now.
Two partial plans are being considered: riding a commuter train or taking a taxi. 
The train leaves in six minutes and
takes 22 minutes to reach the airport.
The planner has not yet
determined how to get to
terminal C:
it may require an additional ten minutes using a bus (with  probability 0.2), or the terminals may be adjacent, requiring no transit time.
The taxi plan entails calling a taxi (2 minutes),
then a ride taking 20 minutes
to get to airport terminal C, and finally a payment step,
the type and length of which the planner has not yet
determined (say one or ten minutes,
each with probability 0.5).
Suppose the remaining planning time for the train plan is known to be eight minutes with 
certainty, and
for the taxi plan it is distributed:
$[0.5:4;0.5:8]$.

This scenario is modeled in
S(AE)$^2$ as follows.
We have two processes: 
process 1 for  the
plan with the commuter train
with $m_1=[1:8]$
and process 2 for the taxi plan
with $m_2=[0.5:4;0.5:8]$,
where we show the PMF $m_i$
rather than the CDF $M_i$ for clarity.
The deadline distribution PMFs are:
$d_1=[0.2:-1,0.8:6]$:  fail with probability 0.2 (the train plus bus plan arrives at terminal C at time 38, which does not meet our goal of 30) and six minutes with probability 0.8 (the train only plan) and $d_2=[0.5:-1,0.5:7]$.

Allocating planning time equally fails with certainty: neither process terminates in time to act. The optimal S(AE)$^2$ policy
realizes that process 1
cannot deliver a
timely solution ($d_1$ is less than $m_1$ with probability 1) and allocates all processing time to process 2, hoping it will terminate
in 4 minutes and find a payment plan that takes only one minute, resulting in success with probability $0.5\times 0.5 = 0.25$.
\end{example}

\subsection{Greedy Schemes}

As solving the metareasoning problem is NP-hard,
\citet{aaai19paper} and \citet{icaps-21} used insights from a diminishing returns result
to develop greedy schemes.
Their analysis is restricted to linear contiguous allocation policies: schedules where the action taken at time $t$ does not depend on the results of the previous actions, and where each process receives its allocated time contiguously.

Following their notation, we denote the probability that process $i$ finds a timely plan when allocated $t_i$ consecutive time units
{\bf b}eginning at time $t_{b_i}$ as:
\begin{equation}
  s_i(t_i,t_{b_i}) = \sum_{t'=0}^{t_i} (M_i(t')-M_i(t'-1)) 
  (1-D_i(t'+t_{b_i}))
\end{equation}

For linear contiguous policies,
we need to allocate  $t_i$, $t_{b_i}$ pairs to
all processes (with no allocation overlap).
Overall, a timely plan is found if
at least one process succeeds, that is, overall failure
occurs only if all processes fail.  Therefore, in order to maximize the probability of overall success (over all possible linear contiguous allocations), we need to allocate  $t_i$, $t_{b_i}$ pairs
so as to maximize the probability:
\begin{equation}\label{eq:success}
  P_s = 1- \prod_i (1-s_i(t_i,t_{b_i})) 
\end{equation}
Using $LPF_i(\cdot)$ (`logarithm of probability of failure') as shorthand for $log(1-s_i(\cdot))$, we note that $P_s$ is maximized if the sum of the $LPF_i(t_i, t_{b_i})$ is minimized and that $-LPF_i(t_i, t_{b_i})$ behaves like a utility that we
need to maximize. For known deadlines, we can assume
that no policy will allocate processing time after
the respective deadline.  We will use $LPF_i(t)$ as shorthand for $LPF_i(t,0)$.

To bypass the problem of non-diminishing returns,
the notion of {\em most effective computation time}
for process $i$ under the assumption that it
begins at time $t_b$ and runs for
$t$ time units was defined as:
\begin{equation}\label{eq:effective}
  e_i(t_b) = \arg\!\min_t \frac{LPF_i(t,t_b)}{t}
\end{equation}
We use $e_i$ to denote $e_i(0)$ below.

Since not all processes can start at time 0,
the intuition from the diminishing returns optimization
is to prefer process $i$ that has the best
utility per time unit, i.e. such that $-LPF_i(e_i)/e_i$ is greatest.
But allocating time to process $i$
delays other processes, so it is also important to
allocate the time now to processes that  have an early deadline. 
\citet{aaai19paper} therefore suggested the following greedy algorithm (denoted as basic greedy scheme, or bgs for short):
Iteratively allocate $t_u$ units of computation time to process $i$ that maximizes:
\begin{equation}\label{eq:greedy}
     Q(i) = \frac{\alpha}{E[D_i]} - 
     \frac{LPF_i(e_i)}{e_i} 
\end{equation}
where $\alpha$ and $t_u$ are positive empirically determined parameters, and $E[D_i]$ is the expectation of the 
random variable that has the CDF $D_i$ (a slight abuse of notation).
The $\alpha$ parameter trades off between preferring
earlier expected deadlines (large $\alpha $) and
better performance slopes (small $\alpha $).

The first part of Equation \ref{eq:greedy} is a somewhat ad-hoc measure of urgency that performs poorly if the deadline distribution has a high variance.
A more precise notion of urgency was defined by \citet{icaps-21} as the damage caused to a process if its computation is delayed by some time $t_u$.
This is based on the available utility gain after the delay of $t_u$.  An empirically determined 
constant multiplier $\gamma$ was used to balance between exploiting the current process reward from allocating time to process $i$ now and the loss in reward due to delay.
Thus, the delay-damage aware (DDA) greedy scheme
was to assign, at each processing allocation round,
$t_u$  time to the process $i$ that maximizes:
\begin{align}\label{eq:impgreedy}
Q'(i) &= \frac{\gamma \cdot LPF_i(e_i(t_u), t_u)}{e_i(t_u)}- 
\frac{LPF_i(e_i, 0)}{e_i}  
\end{align}
The DDA scheme was then adapted and integrated into the OPTIC temporal planner~\cite{DBLP:conf/aips/BentonCC12}, providing state-of-the-art results on situated problems where external deadlines play a significant role \cite{icaps-21}.

\subsection{DP Solution for Known Deadlines}

For S(AE)$^2$ problems with known deadlines (denoted KDS(AE)$^2$),
it suffices to examine linear contiguous policies
sorted by an increasing order of deadlines \cite{aaai19paper}, formally:
\begin{theorem} \label{th:lcp-ordered}
Given a KDS(AE)$^2$ problem, there exists
a linear contiguous schedule with processes sorted by a non-decreasing order of deadlines that is optimal.
\end{theorem}
\noindent Theorem \ref{th:lcp-ordered} was used by \citepw{icaps-21} to obtain a dynamic programming (DP) scheme:
\begin{theorem}\label{th:poly}
For known deadlines, DP according to
\begin{equation}\label{eq:dp}
 \!   OPT(t,l) = \! \max\limits_{0 \leq j \leq d_l - t}\left(\! OPT(t\!+\!j,l\!+\!1)\! -\!LPF_l(j)\right)
\end{equation}
finds the optimal schedule in time polynomial in $n$,
$d_{n}$.
\end{theorem}
\noindent For explicit $M_i$ representations, 
evaluating Equation \ref{eq:dp} in descending order of deadlines
runs in polynomial time.

\section{Concurrent Planning and Execution}\label{sec:defs}

Our new CoPE model extends the
abstract $S(AE)^2$ model to account for
possible execution of actions during search.
We associate each process
with a sequence
of actions, representing the prefix
of a possible complete plan below the node
the process represents. For each process,
there is a plan remainder 
that is still unknown.
In the context of temporal planning, 
these assumptions are reasonable if we
equate each process
with a node in the open list of a forward-search algorithm
that searches from the initial state to the goal
and adds an action when a node is expanded.
Here, the prefix is simply
the list of operators leading to the current node.
The rest of the action sequence is the remaining
solution that may be developed in the future from
each such node. However, here too we
will abstract away from the actual search and model future
search results by distributions.
Thus, in addition to
the $M_i$ distributions over completion
times,
for each process $i$ we have a unique plan prefix $H_i$ ($H$ for head), 
containing a sequence of actions from a set of available
base-level actions $B$. Each action $b\in B$
also has a deadline $D(b)$.
Upon termination, a process $i$ delivers the rest
of the action sequence $\beta_i$ of the solution. As $\beta_i$
is unknown prior to termination,
we model  $dur(\beta_i)$, the duration of $\beta_i$, by a random variable ${\cal R}_i$ whose value
becomes known upon termination.

Unlike in previous work on situated temporal planning, which requires a complete plan before any action is executed, here actions from any action sequence $H_i$
may be executed (in sequence)
even before having a complete plan.
Execution changes the state of the system and we adjust the set of processes to reflect this: any process $i$ where the already
executed action sequence is not a prefix of $H_i$ becomes invalid.
An execution of any prefix of actions from any $H_i$ is legal if and only if: i) the first action starts no earlier than time 0 (representing the current time),
ii) the next action in the sequence begins
at or after the previous action terminates, and iii) every action is executed before its deadline.
Each suffix $\beta_i$
is assumed to be composed of actions
that cannot be executed before process $i$
terminates. Thus $start(\beta_i)$, the time at which $\beta_i$ may begin execution, must be no
earlier than the time at which process $i$ terminates.
We also assume that base-level actions are  non-preemptible and cannot be run in parallel.
However, computation may proceed freely while executing a base-level action.

As in $S(AE)^2$, we have a deadline
for each process,
but with a different semantics:
In $S(AE)^2$, a complete plan must be found in order to start execution. As a result, the deadline distribution in $S(AE)^2$ is defined over the latest possible time to start executing the entire plan.  However, here the requirement is that the execution terminates before the (possibly unknown) deadline;
we call a sequence of actions fully executed before
its deadline a timely execution. Thus, the deadline distribution for CoPE is defined differently.
We begin by assuming that there is a known
distribution (of a random variable ${\cal X}_i$) over $deadline_i$, the deadline
for process $i$,
and again that its true value
becomes known only once the search in process $i$ terminates.
An execution of a solution
delivered by process $i$ is timely
just when 
the  $start(\beta_i)$
occurs in time to
conclude before the process $i$ deadline;
i.e.
$start(\beta_i) \leq  {\cal X}_i -{\cal R}_i$.
We call this value
constraining
$start(\beta_i)$
the {\em  induced deadline}
for process $i$,
and denote it by ${\cal D}_i$.
Note that ${\cal D}_i={\cal X}_i-{\cal R}_i$
is well defined even if ${\cal R}_i$ and ${\cal X}_i$ are dependent.
Thus, we will henceforth simply assume
that the induced deadline ${\cal D}_i$ has a known distribution 
given by its CDF $D_i$ and we can ignore ${\cal X}_i$ and ${\cal R}_i$. So finally, for a process $i$ to be timely,
it must meet two conditions: 1) complete its computation, and 2) complete execution of its entire action prefix $H_i$ before the induced deadline ${\cal D}_i$. In summary, we have:

\begin{definition}
A Concurrent Planning
and Execution problem (CoPE) is, given a set of 
base-level actions $B$ where each
action $b\in B$ has duration $dur(b)>0$, $n$ processes,
each with a (possibly empty) 
sequence $H_i$ of actions from $B$, a performance profile $M_i$, and the induced deadline distribution of each
${\cal D}_i$, to find a policy for allocating
computation time to the $n$ processes
and legally executing base-level actions from some $H_i$,
such that the probability of executing
a timely solution is maximal.
\end{definition}

\begin{example}
Representing the scenario of example 1 (getting to terminal C in 30 minutes) in CoPE, we again have
2 processes with the same performance profiles as in S(AE)$^2$: $m_1=[1:8]$, $m_2=[0.5:4;0.5:8]$.
We also have
$H_1=[\mbox{ride train}]$ and
$H_2=[\mbox{phone, take taxi}]$,
with $dur(\mbox{phone})=2$, $dur(\mbox{ride train})=22$, $dur(\mbox{take taxi})=20$, and the train leaves in six minutes.
The remainder durations are distributed as follows. For $\beta_1$ we have
${\cal R}_1 \sim [0.8:0~;0.2:10]$, and for $\beta_2$ we have ${\cal R}_2 \sim [0.5:1~;~0.5:10]$.
The deadlines are certain in this case,
${\cal X}_1={\cal X}_2=30$, and
the induced deadlines are thus distributed:
${\cal D}_1 \sim [0.8:30~;0.2:20]$ and
${\cal D}_2 \sim [0.5:29~;0.5:20]$.

The optimal CoPE policy here is to run process 2 for four minutes. If it terminates and reveals that ${\cal D}_2=29$ then call for a taxi and proceed (successfully) with the taxi plan. Otherwise (process 2 does not terminate, or terminates and reveals that ${\cal D}_2=20$),
start executing the action from $H_1$: take the train and run process 1, hoping to find
that ${\cal D}_1=30$ (the train stops at terminal C). This policy works with probability of success $P_S=0.25+0.75*0.8=0.85$, as opposed to a success probability of $0.25$ for the optimal solution under S(AE)$^2$. Furthermore, had the terminal C arrival deadline been 25 minutes, all S(AE)$^2$ solutions would have had zero probability of success, while in the CoPE model it is possible to commit to taking the taxi even before planning is done, resulting in a probability of success $P_S=0.5$ (due to the yet unknown payment method).
\end{example}

We have seen how the added complexity of CoPE can pay off by enabling execution concurrently with planning.  However, the problem setting remains well-defined and amenable to analysis.
To analyze CoPE, we make four simplifying assumptions:
1) Time is discrete 
2) The action durations $dur(b)$ are known for all $b\in B$.
3) The variables with distributions $D_i$, $M_i$ are all mutually independent.
4) The individual action deadlines $D(b)$ are irrelevant (not used, or equivalently set to be infinite), as the processes' overall induced deadline distributions $D_i$ are given.
Although assumption 4 is easy to relax (our implementation allows for individual action deadlines), doing so complicates the analysis.

\subsection{Formulating CoPE as an MDP}\label{sec:MDP}
	
We can state the CoPE optimization problem as the solution to an
MDP similar to the one defined for $S(AE)^2$.
The actions in the CoPE MDP are of two types: the actions $c_i$
that allocate the next time unit of computation to process $i$
as in $S(AE)^2$, to which
CoPE adds the possibility to execute a base-level
action from $B$.
We assume that $c_i$ can only
be done if process $i$ has not already terminated
and has not become invalid (and thus fails) by execution of incompatible
base-level actions. An action $b$ from $B$
can only be done when no other base-level action is currently
executing and $b$ is the next action in some
$H_i$ (after the common prefix of base-level actions that all remaining processes share).

The \textbf{states} of the MDP are defined as the
cross product of the following state
variables:
\begin{enumerate*}[label=(\roman*)]
    \item  wall-clock (real) time $T$;
    \item the time $T_i$ already used by process $i$, for all $1 \le i \le n$;

 \item for each process $i$, an indicator of whether it has failed;
\item the time left $W$ until the current base-level action completes execution; and
\item the number $L$ of base-level actions already
      initiated or completed.
 \end{enumerate*}
We also have special terminal states SUCCESS (denoting having found and an executable timely plan) and FAIL (no longer possible to execute a timely plan).	
\noindent The identity of the base-level actions already executed is not explicit in the state, but can be recovered
as the first $L$ actions in any prefix $H_i$, for a process $i$ not already failed.
The initial state $S_0$ has elapsed wall-clock time $T=0$, no
computation time used for any process, so $T_i=0$ for all $1 \le i \le n$, and no base-level actions executed or started so $W=0$ and $L=0$.
The \textbf{reward} function is 0 for all states, except SUCCESS, which has a reward of 1.

The \textbf{transition distribution} is straightforward
(see appendix): deterministic
except for computation actions
$c_i$, which advance the wall-clock time,
and additionally
process $i$ terminates with probability 
$P_{C,i}=\frac{m_i(T_i[S]+1)}{1-M_i(T_i[S])}$
where $T_i[S]$ is the
time already allocated to process $i$ before the current state $S$.
Termination results in 
SUCCESS if the resulting plan
execution can
terminate before the
(now known) deadline, otherwise the process is called failed. 
A process for which there is no
chance for success is called
{\em tardy}. We allow a computation action
$c_i$ only for processes $i$ that have not failed and are not tardy at $S$.
If all processes are either tardy or failed, we transition to the global FAIL state.

\section{Known Induced Deadline CoPE}

Any instance of $S(AE)^2$ can be
made a CoPE instance by setting all $H_i$
to null. Thus finding the optimal solution to CoPE is also NP-hard, even under assumptions 1-4 and
known induced deadlines $d_i$. We denote this
known-induced-deadline restriction of CoPE by KID-CoPE and analyze
this case, trying to get a pseudo-polynomial
time algorithm for computing the optimal policy.

We can represent a policy as an and-tree rooted
at the initial state $S_0$, with each of the agent's actions
as an edge from each state node, leading to a chance node with 
next possible states as children.
A policy tree in which every chance node has
at most one non-terminal child with
non-zero probability is called linear,
because it is equivalent to a simple sequence of
meta-level and base-level actions. 
\begin{lemma}	\label{optimal_schedule}
In KID-CoPE all policies are linear. 
\end{lemma}
\noindent The Proof of Lemma \ref{optimal_schedule} can be found in the appendix.

In KID-CoPE it  thus suffices to find the best linear policy, 
represented henceforth as a sequence $\sigma$
of the actions (both computational and base-level) to be done
starting from the initial state and ending in a terminal state.

Denote by $CA(\sigma)$ the sub-sequence of $\sigma$ that contains just the computation actions. Likewise, $BA(\sigma)$ denotes the base-level actions.
We call a linear policy contiguous if the computation actions for every process are all in contiguous blocks:
\begin{definition}
Linear policy $\sigma$ is contiguous iff $CA(\sigma)[k_1]=CA(\sigma)[k_2]=c_i$ implies
$CA(\sigma)[m]=c_i$ for all $k_1<m<k_2$
and all computation actions $c_i$.
\end{definition}
\begin{theorem}		\label{contiguity}
In KID-CoPE,
there exists an optimal policy that is linear and contiguous.
\end{theorem}
\noindent The proof of Theorem \ref{contiguity} can be found in the appendix.

We still need to schedule the base-level actions.
We show that schedules we call
{\em lazy} are non-dominated.
Intuitively, a lazy policy is one where execution of base-level actions is delayed as long as possible without making the policy tardy or illegal
(e.g., base-level actions overlapping).
Denote by $\sigma_{i \leftrightarrow j}$ the sequence resulting from exchanging the $i$th and $j$th actions in $\sigma$.

\begin{definition}
A linear policy $\sigma$ is lazy if $\sigma_{i \leftrightarrow i+1}$ is tardy
or illegal for all $i$ where $\sigma[i] \in B$.
\end{definition}

\begin{theorem}		\label{laziness}
In KID-CoPE, there exists an optimal policy that is linear, contiguous, and lazy.
\end{theorem}
\noindent The proof of Theorem \ref{laziness} can be found in the appendix.

\section{Pseudo-Polynomial Time Algorithms}\label{sec:pseudo}

For KDS(AE)$^2$, there exists an optimal
linear contiguous policy that assigns computations in order of deadline.
Unfortunately, things are not so simple for
CoPE because the timing of the base-level actions
affects the order in which computation actions
become tardy. 
However, if we fix the
base-level execution start times 
we can then reduce
the resulting KID-CoPE instance into a
KDS(AE)$^2$ instance, which can
be solved in pseudo-polynomial time using DP, as follows.

First, observe that the sequences of
actions we need to consider are only the
$H_i$, as any policy containing an action not in such a sequence
would invalidate all the processes and thus
is dominated. We call the mapping from
$H_i$ to the action execution start times
an {\em initiation function} for $H_i$, denoted by $I_i$.
Note  that a policy $\sigma$ 
with $BA(\sigma)=H_i$ may have computations from other processes $j$, up until such time as $j$ is invalidated by $\sigma$.  Under
a given initiation function $I_i$,
we can define an effective deadline 
$d_j^{\text{eff}}(I_i)$ for
each process $j$, beyond which
there is no point in allowing
process $j$ to run.  Note that the effective deadline is distinct from the known induced process deadline $d_i$.

To define the effective deadline,
let $k \in H_i$ be the first
index at which prefix $H_j$ becomes incompatible with
$H_i$. Then process $j$ becomes
invalid at time $I_i(k)$.
Also, consider any index $m<k$
at which the prefixes are still
compatible. The last time at which
action $H_i[m]$ may be executed
to achieve the known induced deadline $d_j$ is
$t_{i,m}=d_j-dur(H_j[m..|H_j|])$.
That is, process $j$ becomes
tardy at $t_{i,m}$ unless base-level action $H_i[m]$ is executed before
then.
The effective deadline
$d_j^{\text{eff}}$ for process $j$ is thus:
\begin{equation}\label{eq:eff}
d_j^{\text{eff}}(I_i) = \min_{m < k}(I_i(k),
\{ t_{i,m} : t_{i,m} < I_i(H_i[m])\} )
\end{equation}

\begin{theorem}\label{th:fixed}
Among the set of linear contiguous policies for a specific $H_i$ and initiation function $I_i$,
there exists an optimal policy where
the processes are allocated in order
of non-decreasing effective deadlines.
\end{theorem}
\noindent The proof of Theorem \ref{th:fixed} can be found in the appendix.

Given specific $H_i$ and $I_i$,
the proof of Theorem \ref{th:fixed}
constructs an equivalent KDS(AE)$^2$
instance.
This instance can be solved using DP
to find the optimal computations policy
and its success probability.
Now simply iterating over the $n$ possible
$H_i$ and all possible $I_i$,
and taking the best (highest
probability of success) KDS(AE)$^2$ solution, we get the optimal KID-CoPE policy.
The catch is that in general, the number of initiation functions to consider is exponential. But in some cases,
we can show that the number of such initiation functions we need to consider is small.

\paragraph{Bounded Length Prefixes: }
If the length of all the $H_i$
is bounded by a constant $K$,
we get a pseudo-polynomial time
algorithm, because the number of initiation
functions is at most $|dom(T)|^{|H_i|}$.

To implement this technique, we iterate over all possible initiation functions $I_i$ up to length $K$. If $H_i$ is longer than $K$, we complete $I_i$ using some default scheme, 
thereby giving up on optimality. This is what we do in the K-bounded scheme in the next section.

\paragraph{The equal slack case: }
The difference $sl_i = d_i-dur(H_i)$, the {\em slack} of process $i$, is the maximum time we can delay the
actions in $H_i$ without making process $i$ tardy.
The special case of KID-CoPE 
where the slack of
all processes is equal affords the following
pseudo-polynomial time algorithm.
Let the slack of all processes equal $sl$.
Consider any non-tardy
linear contiguous policy $\sigma$ that has $BA(\sigma)=H_i$.
Then $\sigma$ is lazy only
if $I_i^\sigma(H_i[1])=sl$.
In this case the actions in $H_i$ must be executed
contiguously starting at $sl$, as otherwise $\sigma$
would be tardy.
The resulting unique lazy $I_i$ is thus independent of $\sigma$ (as long as 
$BA(\sigma)=H_i$) and due to Theorem \ref{laziness} is the only one we need to consider for each $H_i$.

To summarize this case, for each $H_i$ set the $d^{\text{eff}}_j(I_i)$ for all processes
assuming the unique lazy $I_i$ using Equation
\ref{eq:eff}, solve the resulting KDS(AE)$^2$
instance, and pick the best of these $n$ solutions.

Note that having a known
deadline entails a known induced deadline in
problems with constant length solutions,
such as CSPs. 
But in general (e.g., the 15-puzzle instances
from our empirical evaluation) $dur(\beta_i)$ is unknown before the solution is found,
thus the induced deadline
is also unknown. It is possible for
process $i$ to find a solution,
only to discover that it 
cannot be executed on time, even
for known deadlines.

\section{Algorithms for the General Case}\label{sec:alg}

While optimal, the pseudo-polynomial time algorithms in Section \ref{sec:pseudo} only apply to known deadlines plus additional assumptions.
We thus propose two categories of suboptimal schemes for the general case:
a) focusing primarily on execution or b) focusing primarily on computation.

{\bf Execution-focused schemes} choose some base-level action initiations.
Effective deadlines are computed for each initiation, 
giving S(AE)$^2$ problem instances.
Computations are allocated with
any S(AE)$^2$ algorithm $A$.  Specifically:

\noindent 1) The \textbf{Max-LET$_A$}
schema
considers the minimal value in the support of $D_i$ for each $i$. (Other methods of fixing the deadline can be used, e.g. taking the expectation). Then, for every process $i$, Max-LET$_A$ fixes the base-level actions to their \textbf{L}atest \textbf{E}xecution-\textbf{T}ime at which every action in the head must be executed (with respect to the known deadline). By fixing the base-level actions to those induced by process $i$, the CoPE problem instance is reduced to an S(AE)$^2$ instance. Then algorithm $A$ is executed on the S(AE)$^2$ instance and returns
a linear policy $\pi_i$ and its success probability.
Looking over each process' $\pi_i$, Max-LET$_A$ chooses the one with the highest success probability.

\noindent 2) \textbf{K-Bounded$_A$}
is similar to Max-LET$_A$ with one difference. Instead of fixing base-level actions only to the latest start-time of every process $i$, K-Bounded$_A$ considers all possible placements for the first $K$ actions. The rest (if any) of the time-allocations are determined using latest start-time.

{\bf Schemes focusing on computations} treat a CoPE instance as if it were an S(AE)$^2$ instance, 
ignoring the base-level actions when allocating computation action(s). These schemes can use
any S(AE)$^2$ algorithm to schedule computations.  For example,
\textbf{Demand-Execution$_A$}
first decides which 
computation $c_i$ should be done in the next time unit using S(AE)$^2$ algorithm $A$, under the assumption that no base-level actions are executed before computation terminates.
Then it checks whether a base-level action $b$ is required for $c_i$ to be non-tardy.
If so, the base-level action $b$ is returned, otherwise $c_i$ is returned.

{\bf Hybrid approaches} are also possible.
We tried \textbf{Monte-Carlo tree search (MCTS)},
which is applicable to any MDP \cite{browne2012survey}.
The MCTS version we used utilizes UCT \cite{DBLP:conf/ecml/KocsisS06},
which applies the UCB1 formula \cite{auer2002finite} for selecting nodes, and a random rollout policy that uses $-LPF$ as a value function for sampled time allocations.

\section{Empirical Evaluation}

\begin{figure}[t]
\hspace*{-5pt}
\includegraphics[width=(0.96\linewidth)]{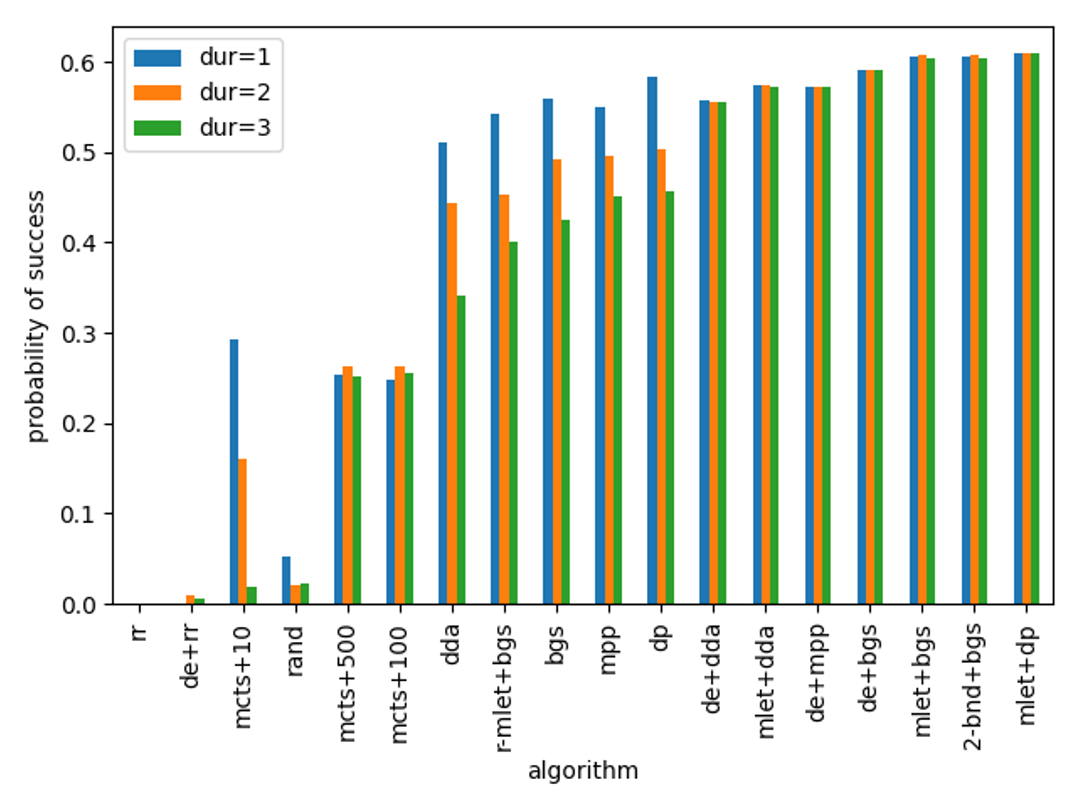}

\vspace*{-18pt}
\includegraphics[width=(0.93\linewidth)]{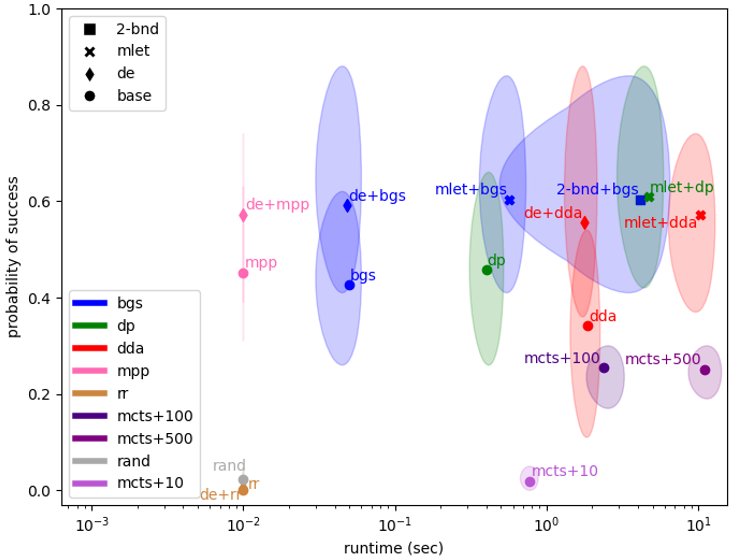}

\vspace{-3mm}

\caption{ \label{fig:main_results}
Results for 20 processes: top) average success probability, bottom) success vs runtime for $dur(b)=3$.}
\end{figure}

Our experimental\footnote{The implementation can be found in the following repository: \url{https://github.com/amihayelboher/CoPE}} setting is inspired by movies such as Indiana Jones or Die Hard in which the hero is required to solve a puzzle before a deadline or suffer extreme consequences.
As the water jugs problem from Die Hard is too easy, we use the 15-puzzle with the Manhattan distance heuristic instead.
We collected data by solving 10,000 15-puzzle instances, recording the number of expansions required by $A^\ast$ to find an optimal solution from each initial state, as well as the actual solution length.
Then, two CDF histograms were created for each initial $h$-value: the required number of expansions, and the optimal solution lengths.
CoPE problem instances of $N$ processes were generated by drawing a random 15-puzzle instance, running $A^\ast$ until the open-list contained at least $N$ search nodes, with $N \in \{2,5,10,20,50\}$, and then choosing the first $N$.
Each open-list node $i$ became a CoPE process, with $M_i$ being the node-expansion CDF histogram corresponding to $h(i)$; ${\cal R}_i$ taken from the solution-cost histogram (to represent the remaining duration of the plan); and $H_i$ the list of actions that leads to $i$ from the start node. In this setting, all base-level actions require the same amount of time units to be completed, denoted as $dur(b)$; in our experiments, we considered $dur(b) \in \{1,2,3\}$ (i.e. each 15-puzzle instance became three CoPE instances, differing only
in the duration of the base-level action). 
Finally, to make the deadlines challenging, we 
used as the deadline for reaching the goal
${\cal X}_i = 4 \times h(i)$. 
Although the value of ${\cal X}_i$ is known, ${\cal D}_i$ is unknown due to ${\cal R}_i$ being unknown.

The empirical evaluation included 10 CoPE instances in each setting.
In order to evaluate the success of the algorithms on each instance, we simulate an outcome by sampling values from the $M_i$ and $D_i$ distribution of each process $i$.
We ran each algorithm 100 times.

From S(AE)$^2$, we implemented the basic greedy scheme (BGS), delay-damage aware (DDA), dynamic programming (DP), round robin (RR, allocate computation time units to all processes cyclically) and most promising plan (MPP, allocate consecutive time to the process with the highest probability to meet the deadline; if the process fails to find a solution, recompute the probabilities with respect to the remaining time).
We implemented a demand-execution version of all the S(AE)$^2$ algorithms (except DP, which is not directly suited to demand-execution), Max-LET$_{\text{DP}}$, Max-LET$_{\text{BGS}}$, Max-LET$_{\text{DDA}}$, 2-bounded$_{\text{BGS}}$ (K-bounded$_{\text{BGS}}$ with $K=2$), and MCTS with an exploration constant $c=\sqrt{2}$ and budgets of $10$, $100$ and $500$ rollouts before selecting each time allocation.

Figure \ref{fig:main_results} shows results for 20 processes.  The top panel show average success probability for base-level $dur(b)$ being 1, 2 and 3.  The results
suggest that most of the algorithms have a high
probability of success for duration=1, corresponding to
a lot of slack.  The performance of the S(AE)$^2$ algorithms in their raw versions (that do not support concurrent execution) gets worse as the time pressure gets higher.
The CoPE-specific algorithms demand-execution, Max-LET and 2-Bounded overcome this problem because they can choose to execute base-level actions while planning, which increases the computation time that can be allocated to processes.
These patterns are repeated for other numbers of processes (see appendix).  Thus, in the following we focus on severe time pressure, $dur(b)=3$.

The bottom panel assesses the trade-off between solution quality (probability of success) and runtime.  As temporal planners such as OPTIC typically expand hundreds of nodes per second, we prefer CoPE algorithms that take less than one second of runtime so that, when eventually integrated into a planner, they could be run every few hundred expansions without too much overhead.  In the plot, ellipse centers are averages over instances, the shaded area just covers the results for all 10 instances.
Max-LET$_{DP}$, Max-LET$_{BGS}$, and 2-Bounded$_{BGS}$ have the best probability of success on average. However, among these schemes
only Max-LET$_{BGS}$ has an acceptable runtime.
The demand-execution schemes exhibit a slightly lower
probability of success but are 
orders of magnitude faster.
The more we take the execution into account, the better the results:
demand-execution does not plan the execution in advance and has the lowest success ratio (among the BGS adaptations to CoPE), Max-LET has a better success ratio as it considers execution first; 2-Bound considers more options than Max-LET, thus has the best success rate among them. However, the additional checks require more computational resources.

The same trends are observed for increasing the number of processes further (see appendix).
Using value iteration to directly solve the MDP delivers the best success probability, but becomes infeasible when the number of processes is $>2$.

\section{Conclusion}\label{sec:con}

Planning is, in general, intractable, so it is unrealistic to assume that time stops during planning.  Starting execution of
a partially developed plan while continuing to search may gain
crucial time to deliberate
at some risk of performing
actions that do not lead to a solution.
We extended the abstract metareasoning
model for situated temporal planning
of \citepw{aaai19paper} to allow for concurrent action execution and deliberation.
Our CoPE problem is NP-hard, but pseudo-polynomial time algorithms are possible
for the cases of bounded-length plan prefixes and of equal slack.
We developed several
suboptimal
algorithms for
the case of unknown deadlines and
suffix durations. Experiments
based on the popular 15-puzzle benchmark showed that the new algorithms span a useful range of trade-offs between runtime and probability of a timely solution.

Now that a formal framework and principled algorithms have been introduced, the obvious next step is to adapt and integrate these schemes into a situated online planner.
Doing so requires online estimation of the necessary statistics and careful engineering to keep the metareasoning overhead manageable.  Extending the model to parallel durative actions is another promising direction.

\section*{Acknowledgements}
This research was supported by grant 2019730 from the United States-Israel Binational Science Foundation (BSF) and grant 2008594 from the United States National Science Foundation (NSF).

\bibliography{main}

\centering\section*{APPENDIX}

\appendix

\section{Formulating CoPE as an MDP}
	
We can state the CoPE optimization problem as the solution to an
MDP similar to the one defined for $S(AE)^2$.
The actions in the MDP are of two types: executing a base-level
action from $B$, and the actions $c_i$
that allocate the next time unit of computation to process $i$.
We assume that $c_i$ can only
be done if process $i$ has not already terminated
and has not become invalid by execution of incompatible
base-level actions. An action $b$ from $B$
can only be done when no base-level action is currently
executing and $b$ is the next action in some
$H_i$ (after the common prefix of base-level actions that all remaining processes share).

The \textbf{states} of the MDP are defined as the
cross product of the following state
variables:
\begin{itemize}
    \item Wall clock (real) time $T$.
    \item Time $T_i$ already assigned to each process $i$, for all $i$ from 1 to $n$.
      These variables are also used to encode process failure
      to find a timely solution, thus
      $dom(T_i)\in \mathbb{N} \cup \{F\}$.
      The value $F$ is also used to indicate any process $i$ 
      with $H_i$ inconsistent with the already executed base-level actions. (This is a compression of these two variables into one state variable, a tweak not mentioned in the body of the paper for clarity.)
\item Time left $W$ until the current base-level
      action completes execution.
\item The number $L$ of base-level actions already
      initiated or completed.
\end{itemize}
We also have special terminal states SUCCESS (denoting having found and an executable timely plan) and FAIL (no longer possible to execute a timely plan).	Thus, the state space of the MDP is:
\begin{eqnarray*}
\mathcal{S} & = & (dom(T) \times dom(W) \times dom(L) \times \\
		~ & ~ &  {\LARGE \bigtimes}_{1 \le i \le n} dom(T_i)) 
		~~~ \cup \{ \mbox{SUCCESS, FAIL} \}
\end{eqnarray*}
\noindent The identity of the base-level actions already executed is not explicit in the state, but can be recovered
as the first $L$ actions in any prefix $H_i$, for a process $i$ not already failed.
The initial state $S_0$ has elapsed wall clock time $T=0$, no
computation time used for any process, so $T_i=0$ for all $1 \le i \le n$, and no base-level actions executed or started so $W=0$ and $L=0$.
The \textbf{reward} function is 0 for all states, except SUCCESS, which has a reward of 1.

The \textbf{transition distribution} is determined by which process $i$ is being scheduled (a $c_i$ action) or how execution has proceeded (a $b$ action). For simplicity
we assume that only one action
is applied at each transition, although
base level and computation actions
can overlap in real (wall clock) 
time.
Let $S = (T,T_1...T_n,W,L)$ be a state and $S'$
be the state after an action is executed.
We use the notation $var[state]$ to denote the value
of state variable $var$ in $state$,
for example $T[S]$ denotes the value of $T$ in $S$,
that is, the value of the wall-clock time in state $S$.
For a base-level action $b\in B$, which is only
allowed if $W[S]=0$, the transition is
deterministic: the count of executed actions increases
and all processes incompatible with $b$ fail. That is,
$W[S']=dur(b)$, $L[S']=L[S]+1$, $T[S']=T[S]$, and:
		\[
		  T_i[S'] =  \left\{ \begin{array}{ll} 
		       T_i[S] & \mbox{if} ~~ |H_i| > L[S] ~\mbox{and}~ H_i[L[S]+1] = b \\
		       F   &  \mbox{otherwise}
		  \end{array}
		  \right.
		\]

In most cases (exception stated below), a computation action advances the wall-clock time.
As a result, some processes may become unable to
deliver a timely solution; we call such processes,
as well as their computation actions, {\em tardy}.  Formally, consider any process $i$ that has still not failed
in state $S$.
Denote $t_i[S]$ the earliest time at which execution of a solution generated by process $i$ can complete.
With $[i..j]$ denoting a sub-sequence from $i$ to $j$, inclusive, and $dur(.)$ of a sequence of actions denoting the sum of durations of the actions in the sequence, we have:
\[
t_i[S] = T[S]+W[S]+dur(H_i[(L[S]+1)..|H_i|])+1
\]
That is, $t_i[S]$ equals time now, plus time remaining until the current base-level action (if any) terminates, plus the duration of the
tail of the $H_i$ prefix, plus the 1 time unit allocated now.
The probability that this is a timely execution
is $1-D_i(t_i[S])$. A process $i$ for which $D_i(t_i[S])=1$
has zero probability of delivering a timely execution and
is called tardy. When doing a computation
action, each process $i$ that is tardy at $S$ fails,
that is, $T_i[S']=F$ with probability 1, unless
all processes are tardy, in which case we fail globally,
i.e. $S'=\mbox{FAIL}$. The above transitions are
deterministic.

We allow a computation action
$c_i$ only for processes $i$ that have not failed and are not tardy at $S$.
For such a valid action $c_i$,
we have $T[S']=T[S]+1$, $W[S']=max\{0, W[S]-1\}$, and $T_j[S']=T_j[S]$ for all $j\neq i$ that are non-tardy.
With probability $P_{C,i}=\frac{m_i(T_i[S]+1)}{1-M_i(T_i[S])}$
process $i$ now terminates, given that it has not terminated before. Thus with probability $1-P_{C,i}$ the process does not terminate, in which case we get $T_i[S']=T_i[S]+1$.
If the process does terminate, as stated above,
it delivers a timely solution
with probability $1-D_i(t_i[S])$ in which case we set $S=\mbox{SUCCESS}$.
The solution fails to meet the induced deadline with probability
$D_i(t_i[S])$, in which case we have $T_i[S']=F$, unless
in the resulting $S'$ there is no longer any non-tardy process that has not failed, in which case set $S'=\mbox{FAIL}$.

\section{Proofs}

\paragraph{Proof of Lemma 3}
\begin{proof}
Transitions for base-level actions are
deterministic, thus it is sufficient to consider deliberation actions $c_i$ at any state $S$. 
Examining the transition distribution here, the only
chance nodes with more than one non-terminal child can occur when
a process $i$ may terminate and fail.
However, since the induced deadlines
are all known, then $D_i(t_i[S])$ is
a step function, i.e. has value
either 0 or 1.
The case $D_i(t_i[S])=1$ means process $i$ is tardy,
so $c_i$ is not allowed, and such chance nodes cannot occur.
\end{proof}

\paragraph{Proof of Theorem 4}
\begin{proof}
From the proof of Lemma 	3, 
and the CoPE MDP definition, an optimal linear KID-CoPE policy is non-tardy and
any process that terminates results
in SUCCESS. Due to independence between the
$M_i$, the probability of termination (and thus
success) of each process depends only on the
total processing time $a_i$ allocated to it
and equals $M_i(a_i)$. Therefore, the total
probability of success is invariant
to the order of computation actions,
as long as all computation actions
do not cause $i$ to become tardy. It is thus sufficient
to show that every linear non-tardy policy
can be re-arranged into one that is contiguous.

Let $\sigma$ be an optimal linear policy
and $k$ be the last index where contiguity is
violated in $CA(\sigma)$. That is, 
the subsequence $CA(\sigma)[(k+1)..|CA(\sigma)|]$
is contiguous, but we have
$CA(\sigma)[k]=c_j$,  
$CA(\sigma)[k+1]=c_i \neq c_j$,
and there exists $m<k$ such that $CA(\sigma[m])=c_i$.
Replacing $CA(\sigma)$
by $CA(\sigma)_{m\leftrightarrow k}$ 
in $\sigma$ still results
in a non-tardy policy because the moved
$c_j$ is made earlier, so cannot become tardy due to this change, and the moved $c_i$
also does not become tardy as there is a later $c_i$ that is non-tardy.
Also,
$CA(\sigma)_{m\leftrightarrow k}[k...|CA(\sigma)|]$
is contiguous by construction.
Such exchanges can be repeated until the policy becomes contiguous.
\end{proof}

\paragraph{Proof of Theorem 5}
\begin{proof}
Define a lexicographic ordering $>_L$ on 
linear policies w.r.t. the index at which their
base-level actions occur: $x >_L y$
if, for some $k\geq 0$, the first $k$ base
level actions in $x$ and
$y$ start at equal indices
and the $k+1$ action of $x$ starts later than
that of $y$.
Let $\sigma$ be the optimal contiguous linear policy that is greatest w.r.t. $>_L$.
Assume in contradiction that $\sigma$ is not lazy.
Then, by definition there exists an index
$i$ such that $\sigma [i]\in B$ and
$\sigma_{i \leftrightarrow i+1}$ is legal and non-tardy and contiguous
(as $CA(\sigma)$ order is unchanged).
Note that $\sigma_{i \leftrightarrow i+1}$
has the same computation time assigned to each process
as $\sigma$, thus being non-tardy, has the same
probability of success as $\sigma$, so
is optimal.
But $\sigma_{i \leftrightarrow i+1} >_L
\sigma$, a contradiction.
\end{proof}

\paragraph{Proof of Theorem 6}
\begin{proof}: For base-level initiation function $I_i$, by construction, process $j$ 
results in a timely execution iff it terminates in time before $d_j^{\text{eff}}(H_i)$.
Thus, linear contiguous policies that have computational actions $c_j$ after $d_j^{\text{eff}}(I_i)$ are dominated.

Setting deadline $d'_j=d_j^{\text{eff}}(I_i)$
we can now ignore the base-level actions, 
getting an equivalent KDS(AE)$^2$ instance
with each $d'_j$ acting as the known deadline
for process $j$. The theorem thus follows
immediately from Theorem 1.
\end{proof}

\section{Pseudo-code of Algorithms}
Algorithms~\ref{alg:max_let},\ref{alg:demand_execution}, and \ref{alg:refined_max_let} show the pseudocode for Max-LET$_A$, Demand-Execution$_{SAE2\_Alg}$ and Refined-Max-LET$_{SAE2\_Alg}$, respectively.
\begin{algorithm}[htb]
\caption{Max-LET$_A$}\label{alg:max_let}
\vspace{0.4cm}
\KwData{IPAE instance $I$, S(AE)$^2$ algorithm $SAE2\_Alg$}
\KwResult{policy $\sigma$}
$\sigma \gets ()$\\
$p \gets 0$\\
\For{$i \in [1,n]$}{
    \vspace{0.2cm}
    $\ d_i \gets min($SUPP$[D_i])$ \\
    $H_i\_sched\ \gets\ schedule\_actions(d_i,\ H_i$) \\
    $S\ \gets\ reduce\_ipae\_to\_sae2(I,\ H_i,\  H_i\_sched)$\\
    $p_i,\ \sigma_i\ \gets\ SAE2\_Alg(S)$ \\
    \If{$p_i > p$}{
        \vspace{0.2cm}
        $p\ \gets\ p_i$ \\
        $\sigma \gets combine\_actions\_in\_\sigma(H_i,\ H_i$\_sched,$\ \sigma_i)$
    }
}
\Return{$\sigma$}
\vspace{0.4cm}
\end{algorithm}

\vspace{0.4cm}
\begin{algorithm}[tb]
\caption{Demand-Execution$_{SAE2\_Alg}$ (Algorithmic Scheme) }\label{alg:demand_execution}
\vspace{0.4cm}
\KwData{state $S$ of IPAE instance, S(AE)$^2$ algorithm ${SAE2\_Alg}$}
\KwResult{base-level action or meta-level action $a \in \{c_i\}_{i=1}^{n} \cup B$}
$i\ \gets\ {SAE2\_Alg}(S)$ \\
$d_i \gets min($SUPP$[D_i])$ \\
$H_i\_sched\ \gets\ schedule\_actions(d_i,\ H_i$) \\
$ind\ \gets\ W[S]$ \\
\eIf{$H_i\_sched[ind] = T[S]$}{
    \vspace{0.2cm}
    $b\ \gets\ H_i[ind]$ \\
    \Return{b}
}{
    \vspace{0.2cm}
    \Return{$c_i$}
}
\vspace{0.4cm}
\end{algorithm}

\vspace{0.4cm}
\begin{algorithm}[tb]
\caption{Refined-Max-LET$_{SAE2\_Alg}$}\label{alg:refined_max_let}
\vspace{0.4cm}
\KwData{IPAE instance $I$, S(AE)$^2$ algorithm ${SAE2\_Alg}$}
\KwResult{policy $\sigma$}
$H \gets ()$ \\
$P \gets \{1...n\}$ \\
\While{$|P| \ne 1$}{
    \vspace{0.2cm}
    // $i$ is a process index, $t_i$ is the time to execute the next action in $H_i$\\
    $i, t_i \gets argmax_{i \in P}{SAE2\_Alg}(reduce\_ipae\_to\_sae2(I,\ H\cup\{H_i[H.length]\}))$\\
    $insert\_to\_the\_end(H,\ (H_i[H.length], t_i))$\\
    $remove\_incompatible\_processes(P,\ i,\ t_i,\ H.legnth)$
}
$p,\ \sigma\ \gets\ {SAE2\_Alg}(reduce\_ipae\_to\_sae2(I,\ H))$\\
\Return{$combine\_actions\_in\_\sigma(H,\ \sigma)$}
\vspace{0.4cm}
\end{algorithm}

\section{Empirical Evaluation: Detailed Results}
\label{app:res}

The experiments were implemented in Python, and run on a computer with an Intel i5-5200U CPU (2.20GHz), with 8GB of RAM and 2 cores, on a Windows 10 (version 1803 64 bit) OS.
This appendix contains all the empirical results from the 15-puzzle based instances, as there was insufficient space to state them in the paper. The typical results all appeared in the paper, so the results here are mostly for completeness, although there are a few minor phenomena that can be observed in these additional results.

For every $N \in \{2, 5, 10, 20, 50\}$, we first display bar charts of the average probability of success and average runtime of all the instances with $N$ processes.
We then display separate results for instances with different durations of base-level actions.

Remark: all the algorithms with runtime less than $10^{-2}$ seconds were rounded to $10^{-2}$ in order to appear in the logarithmic-scaled figures.

The reasons we chose to display the case of instances of 20 processes are as follows.
When the number of processes is greater than 20, the probability of finding a solution is higher.
Therefore, instances with a high number of processes might be not interesting.
Specifically, in many of the instances with 50 processes the success probability was very close to 1, but for $N \le 20$ this was not the case.
On the other hand, when the number of processes was too small, which algorithm was used made little
difference, even in the case where base-level actions take 3 time units.

\begin{figure}[ht]
\begin{center}
\includegraphics[width=0.45\columnwidth]{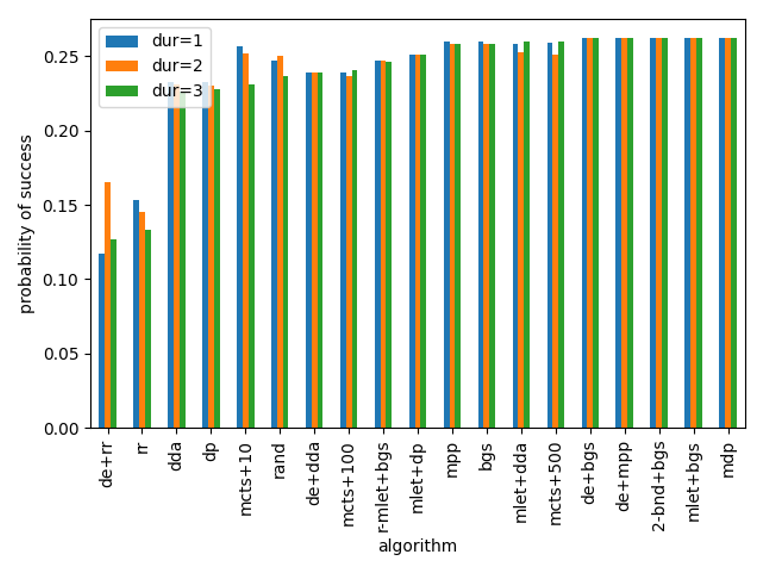}
\includegraphics[width=0.45\columnwidth]{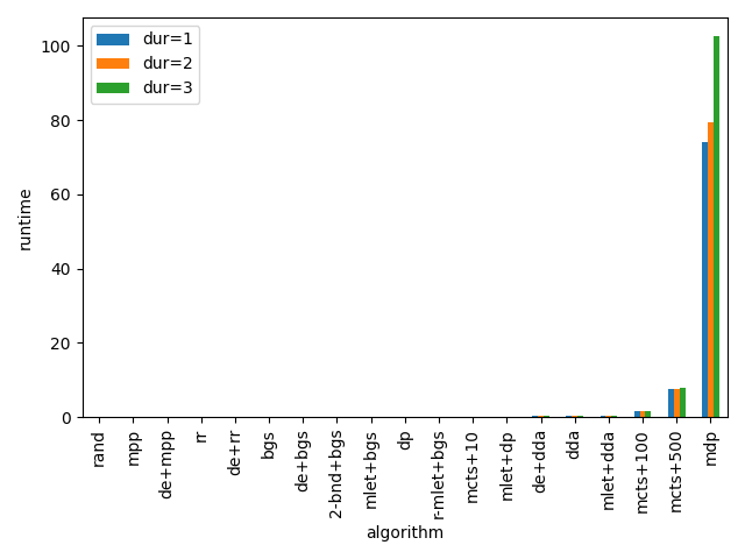}
\end{center}
\caption{Bar plots for 2 processes.}
\label{N=2:bars}
\end{figure}

\begin{figure}[ht]
\begin{center}
\includegraphics[width=0.4\columnwidth]{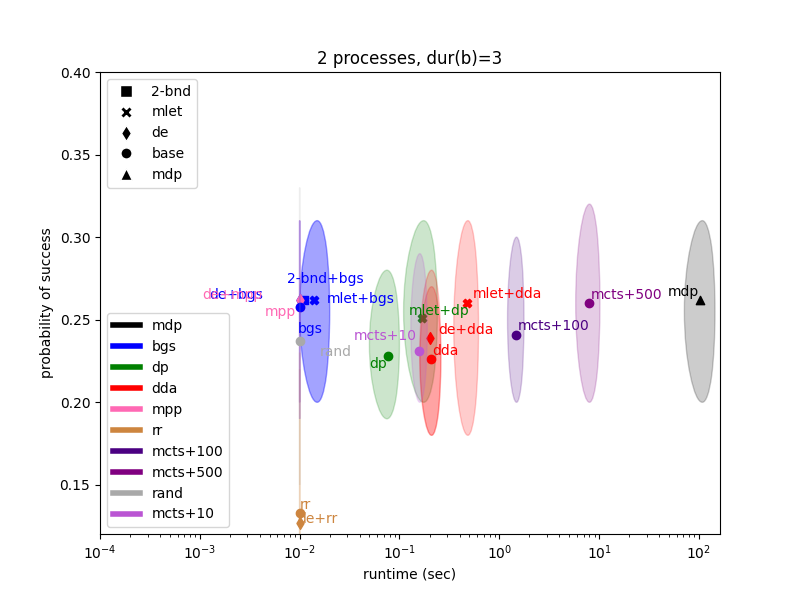}
\end{center}
\caption{Average results for instances with 2 processes and $dur(b)=3$. Ellipse centers are averages over instances, the shaded area just covers the results for all 10 instances.}
\label{N=2:dur=3:ellipses}
\end{figure}

\begin{figure}[ht]
\begin{center}
\includegraphics[width=0.4\columnwidth]{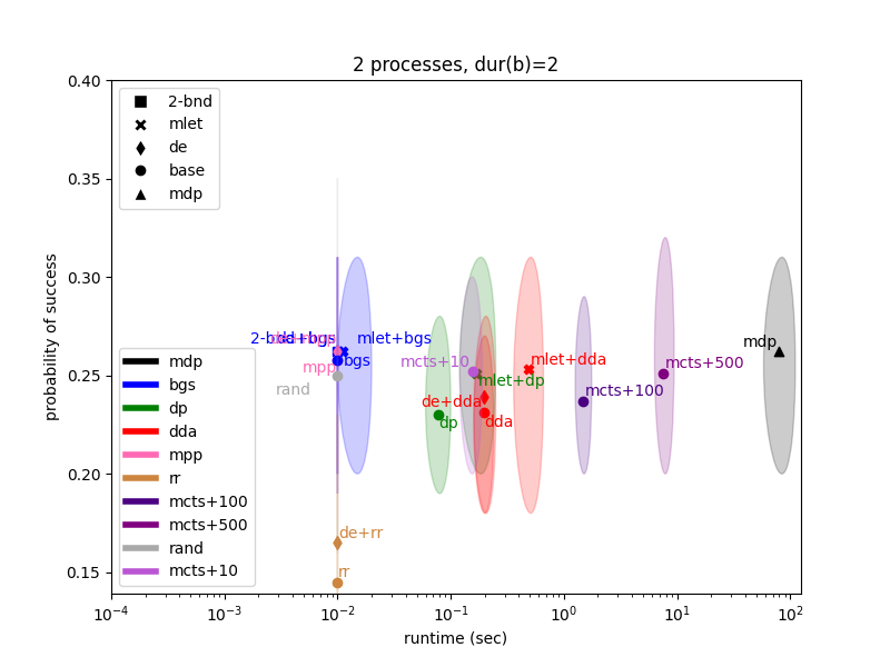}
\end{center}
\caption{Average results for instances with 2 processes and $dur(b)=2$. Ellipse centers are averages over instances, the shaded area just covers the results for all 10 instances.}
\label{N=2:dur=2:ellipses}
\end{figure}

\begin{figure}[ht]
\begin{center}
\includegraphics[width=0.4\columnwidth]{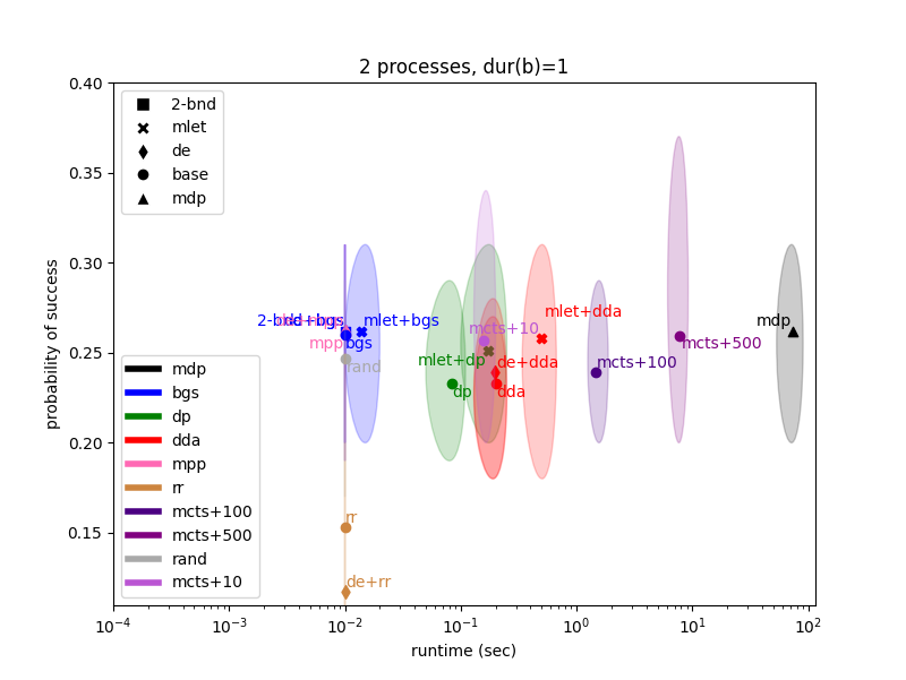}
\end{center}
\caption{Average results for instances with 2 processes and $dur(b)=1$. Ellipse centers are averages over instances, the shaded area just covers the results for all 10 instances.}
\label{N=2:dur=1:ellipses}
\end{figure}

\begin{figure}[ht]
\begin{center}
\includegraphics[width=0.45\columnwidth]{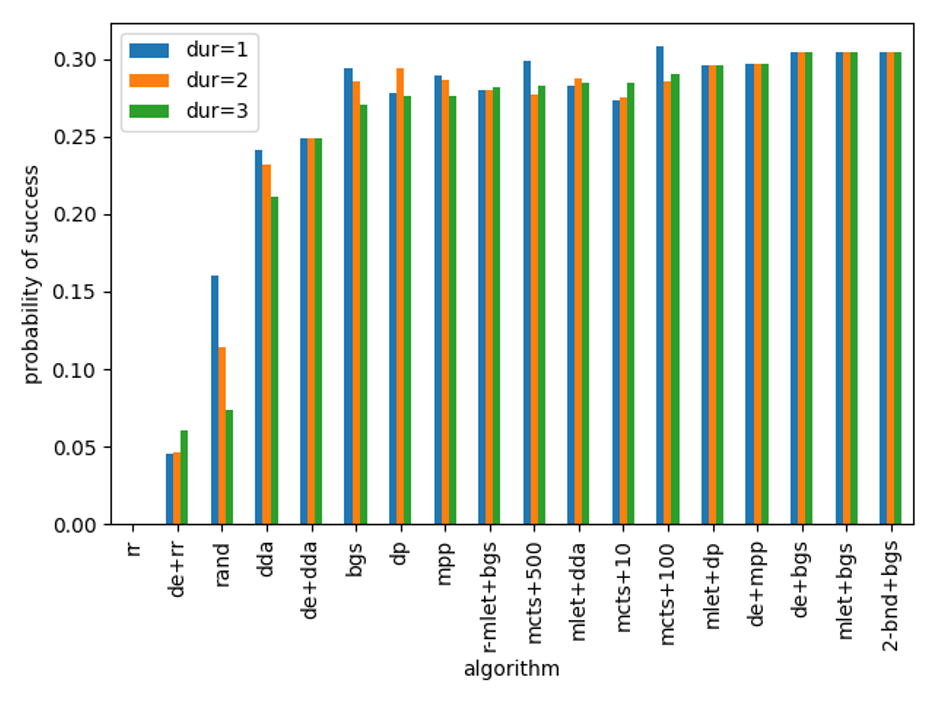}
\includegraphics[width=0.45\columnwidth]{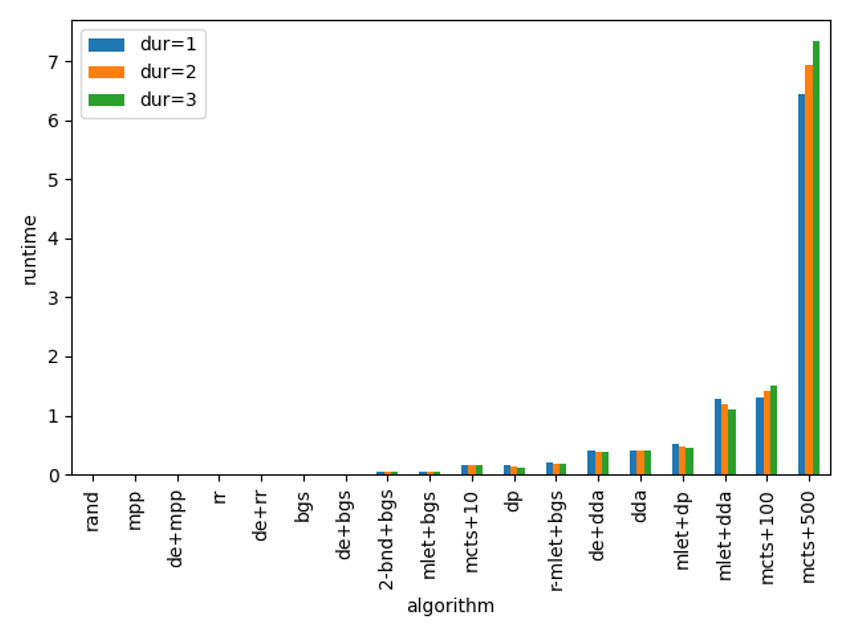}
\end{center}
\caption{Bar plots for 5 processes.}
\label{N=5:bars}
\end{figure}

\begin{figure}[ht]
\begin{center}
\includegraphics[width=0.4\columnwidth]{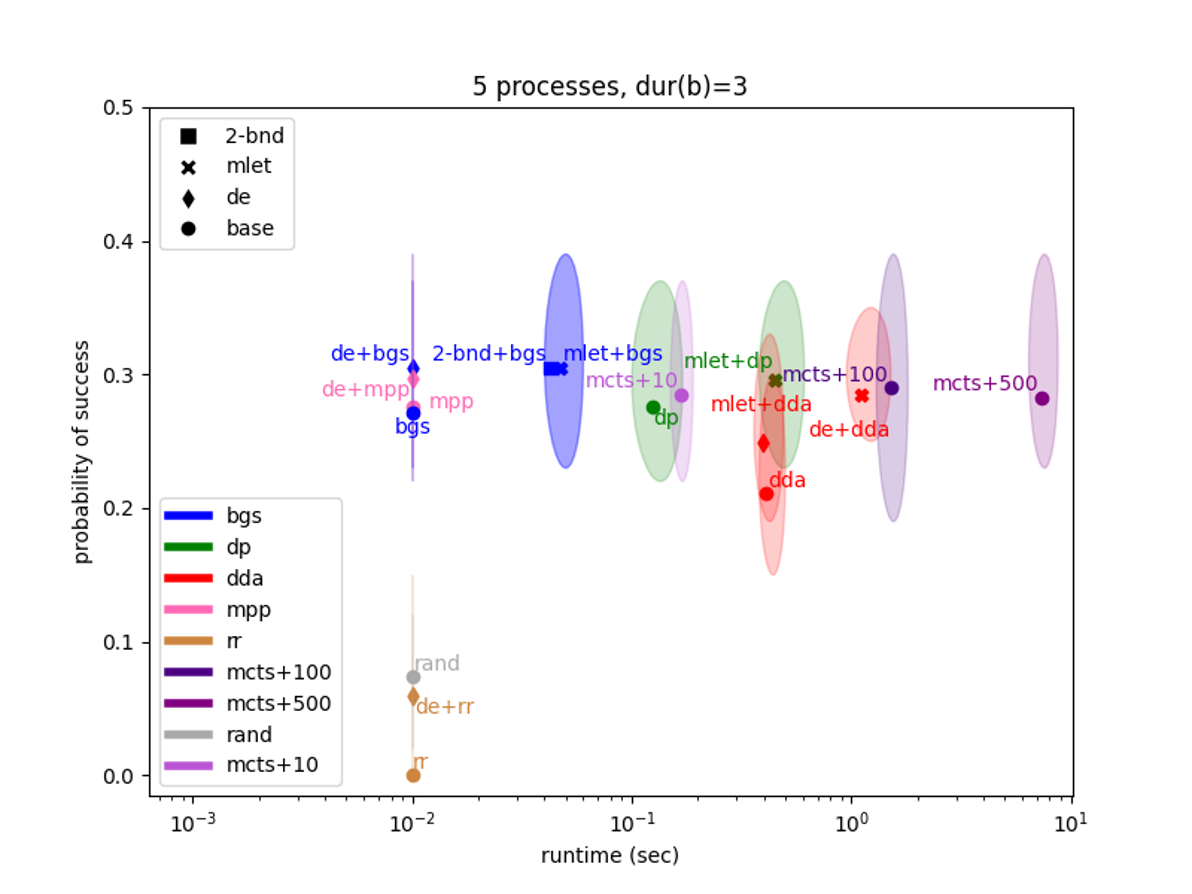}
\end{center}
\caption{Average results for instances with 5 processes and $dur(b)=3$. Ellipse centers are averages over instances, the shaded area just covers the results for all 10 instances.}
\label{N=5:dur=3:ellipses}
\end{figure}

\begin{figure}[ht]
\begin{center}
\includegraphics[width=0.4\columnwidth]{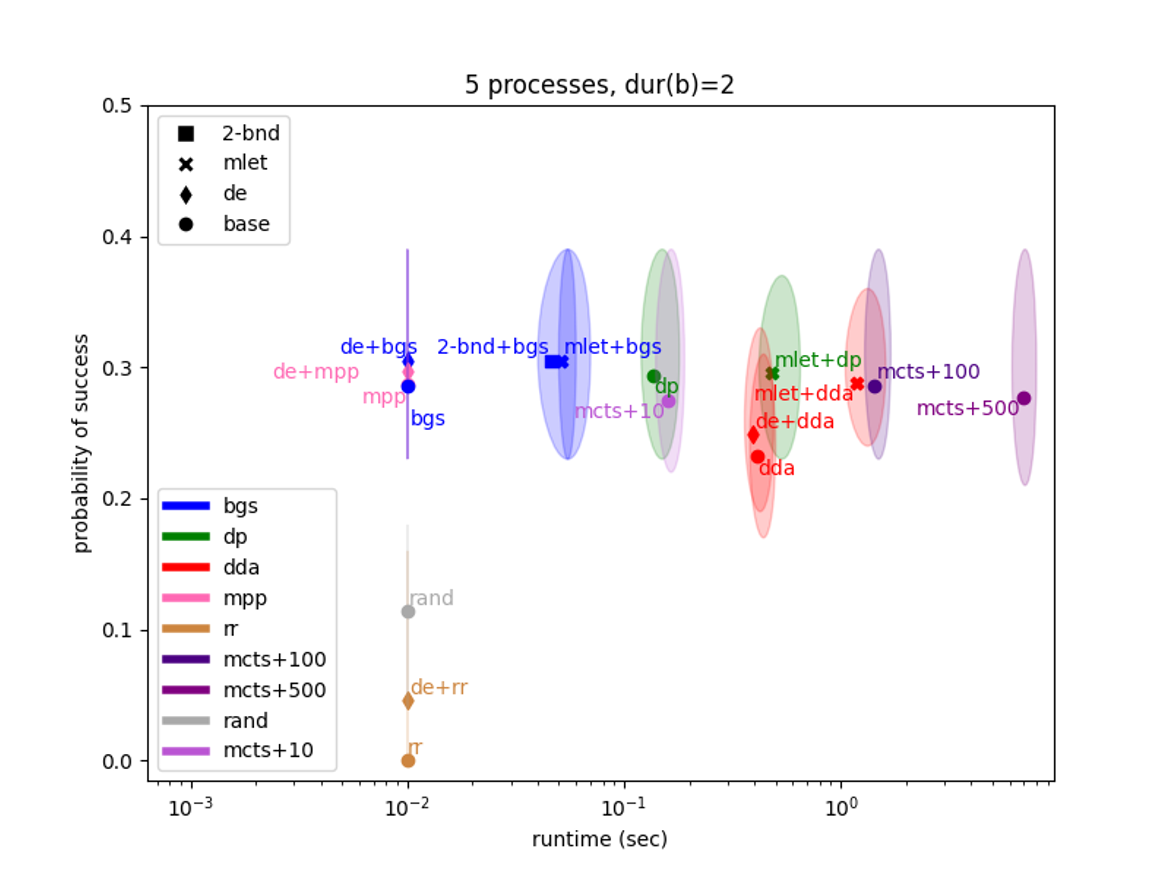}
\end{center}
\caption{Average results for instances with 5 processes and $dur(b)=2$. Ellipse centers are averages over instances, the shaded area just covers the results for all 10 instances.}
\label{N=5:dur=2:ellipses}
\end{figure}

\begin{figure}[ht]
\begin{center}
\includegraphics[width=.4\columnwidth]{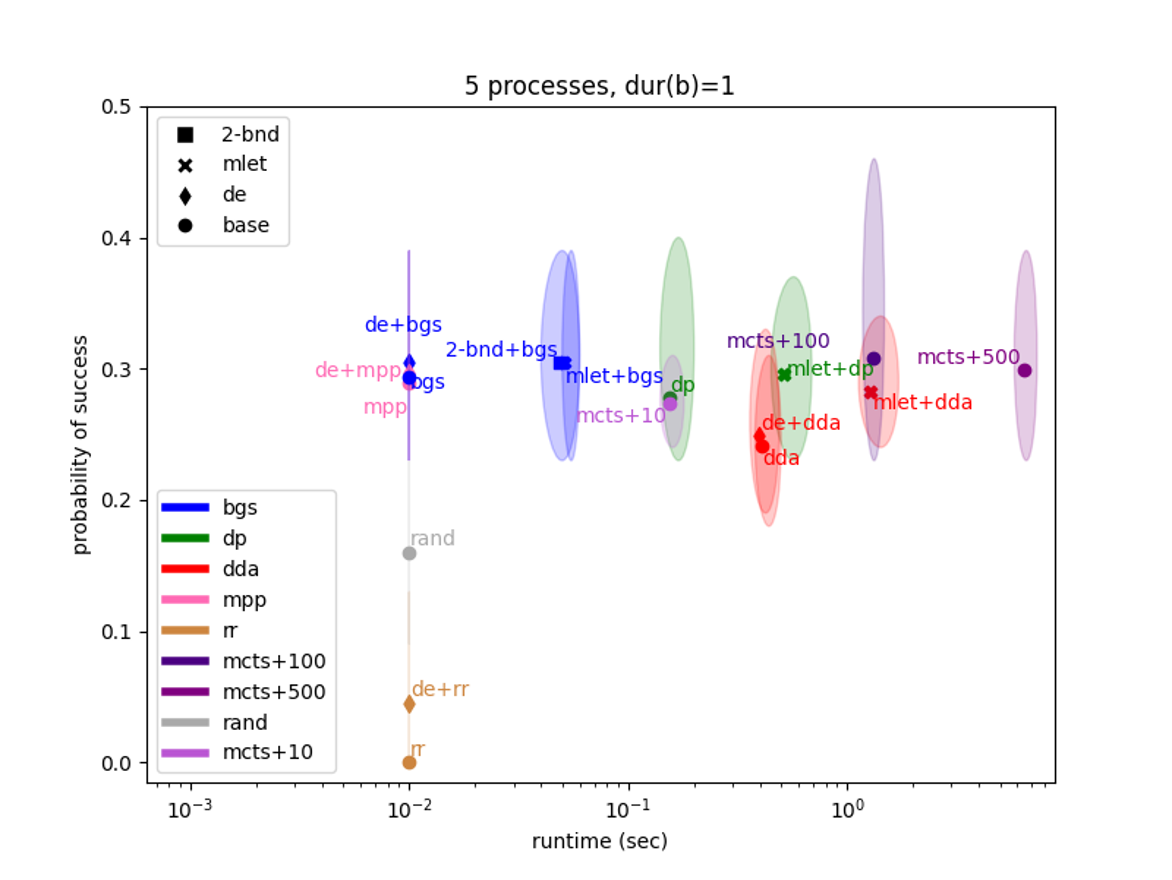}
\end{center}
\caption{Average results for instances with 5 processes and $dur(b)=1$. Ellipse centers are averages over instances, the shaded area just covers the results for all 10 instances.}
\label{N=5:dur=1:ellipses}
\end{figure}

\begin{figure}[ht]
\begin{center}
\includegraphics[width=0.45\columnwidth]{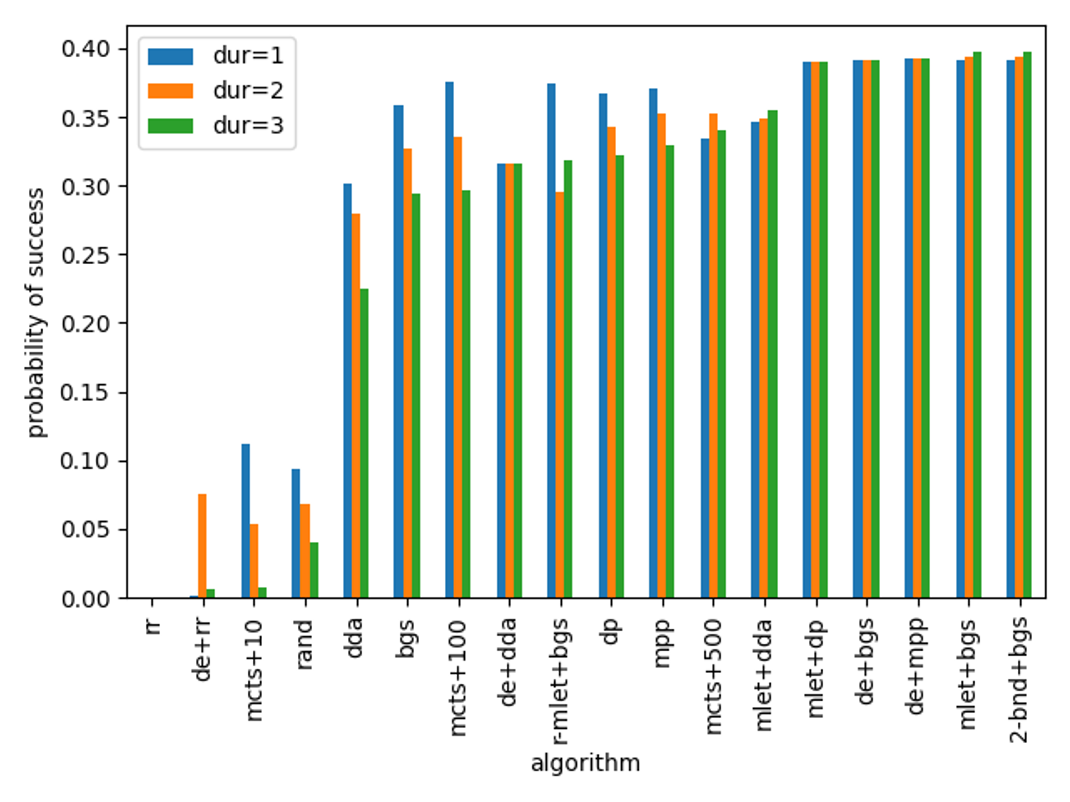}
\includegraphics[width=0.45\columnwidth]{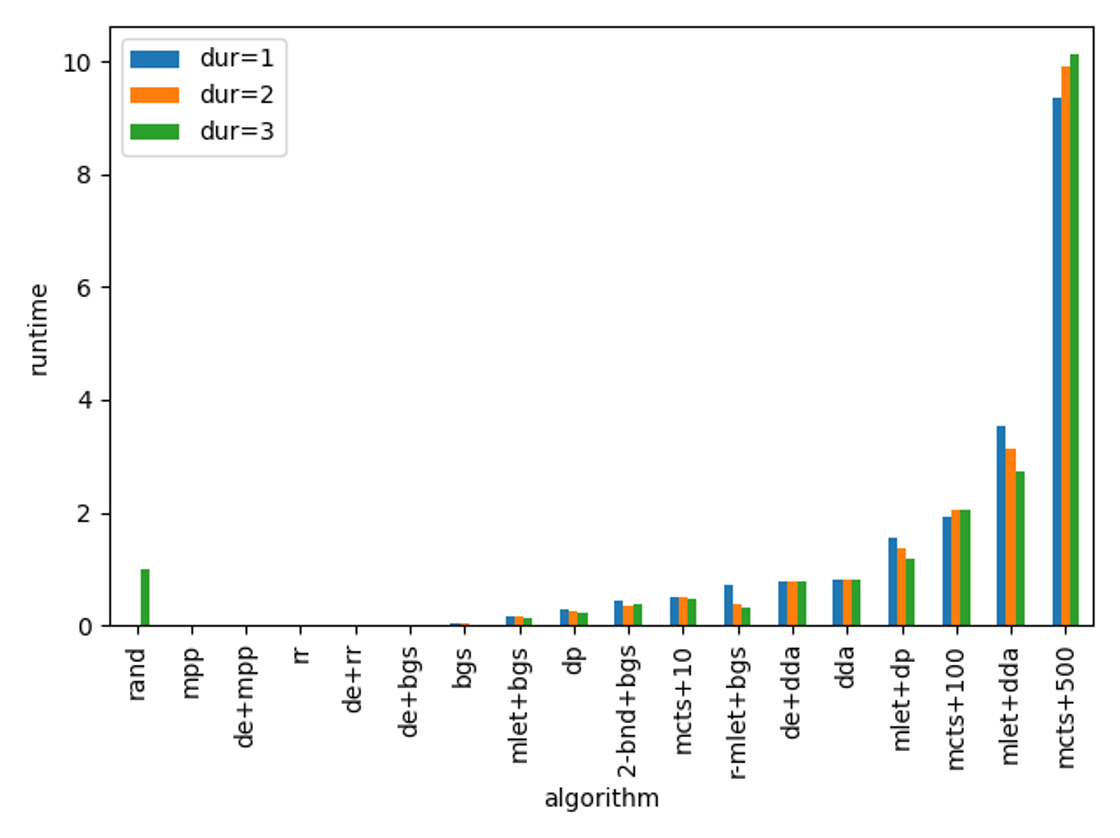}
\end{center}
\caption{Bar plots for 10 processes.}
\label{N=10:bars}
\end{figure}

\begin{figure}[ht]
\begin{center}
\includegraphics[width=0.4\columnwidth]{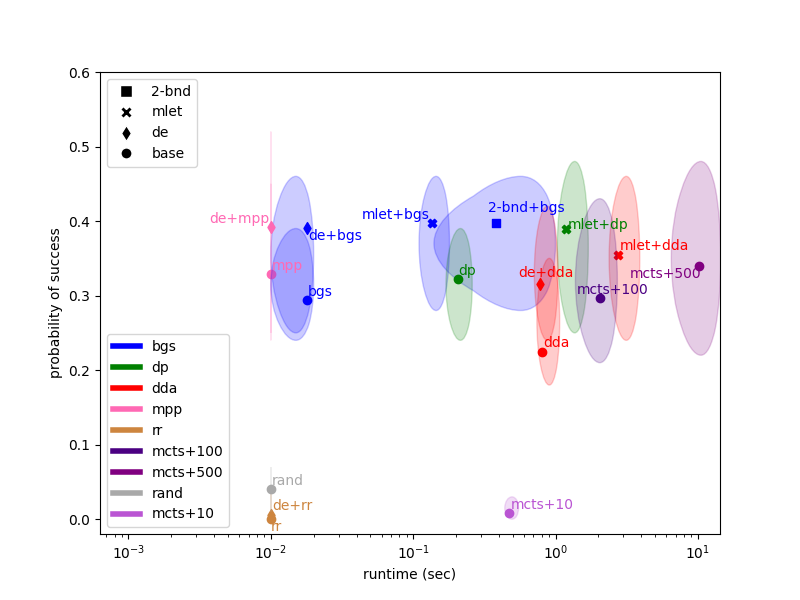}
\end{center}
\caption{Average results for instances with 10 processes and $dur(b)=3$. Ellipse centers are averages over instances, the shaded area just covers the results for all 10 instances.}
\label{N=10:dur=3:ellipses}
\end{figure}

\begin{figure}[ht]
\begin{center}
\includegraphics[width=0.4\columnwidth]{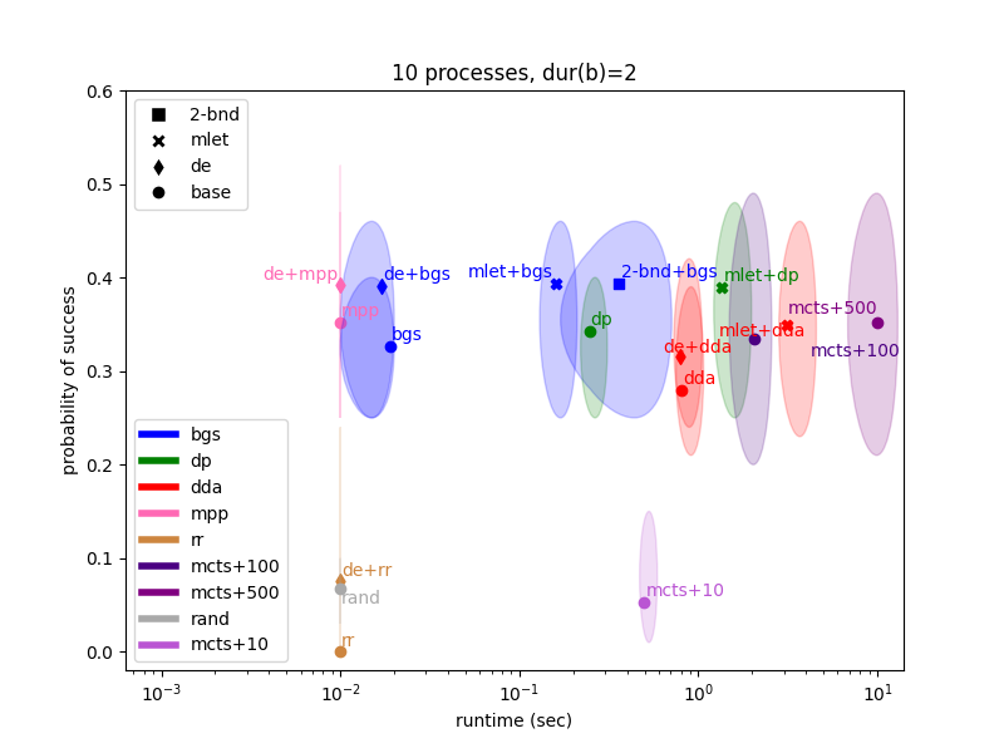}
\end{center}
\caption{Average results for instances with 10 processes and $dur(b)=2$. Ellipse centers are averages over instances, the shaded area just covers the results for all 10 instances.}
\label{N=10:dur=2:ellipses}
\end{figure}

\begin{figure}[ht]
\begin{center}
\includegraphics[width=0.4\columnwidth]{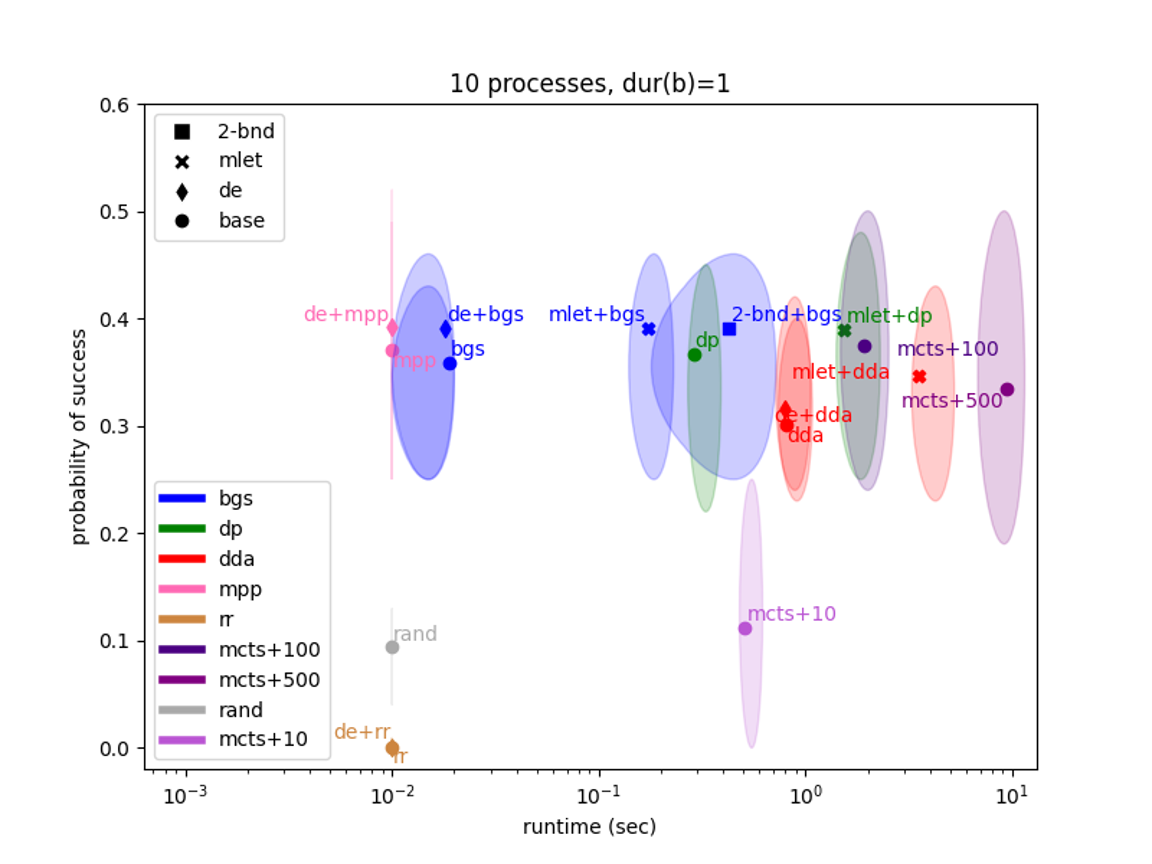}
\end{center}
\caption{Average results for instances with 10 processes and $dur(b)=1$. Ellipse centers are averages over instances, the shaded area just covers the results for all 10 instances.}
\label{N=10:dur=1:ellipses}
\end{figure}

\begin{figure}[ht]
\begin{center}
\includegraphics[width=0.45\columnwidth]{plots/20 processes/bar_plot_success.png}
\includegraphics[width=0.45\columnwidth]{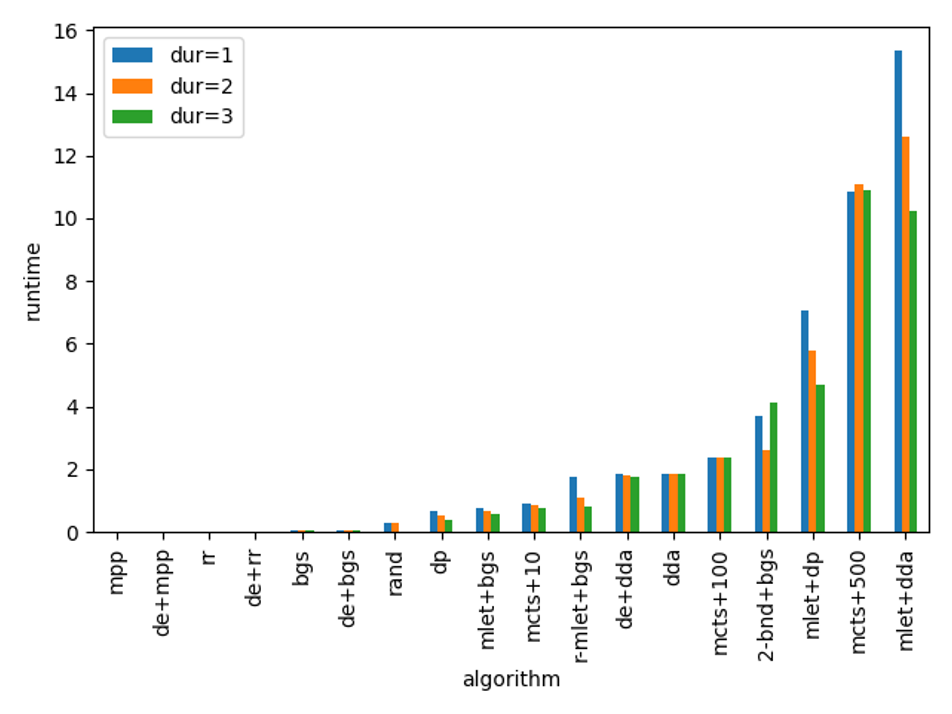}
\end{center}
\caption{Bar plots for 20 processes. In both figures, the blue, orange and green bars represent the results for the instances where the base-level action durations are 1, 2 and 3, respectively. The left figure shows the average probability of success of each algorithm, and is sorted according to the green bars (because the time pressure is highest when the duration is 3). The right figure shows the average runtime of each algorithm, and is sorted according to the blue bars (for aesthetic reasons).}
\label{N=20:bars}
\end{figure}

\begin{figure}[ht]
\begin{center}
\includegraphics[width=0.45\columnwidth]{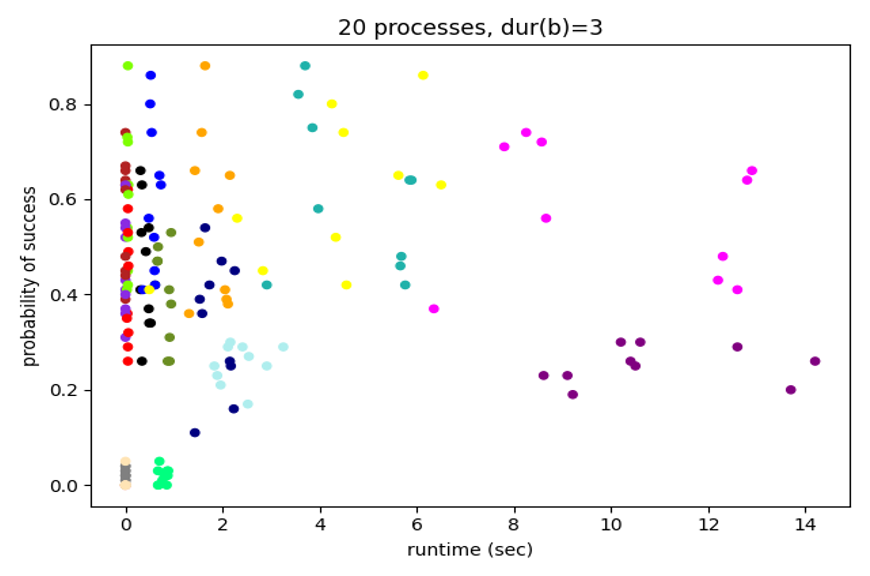}
\includegraphics[width=0.45\columnwidth]{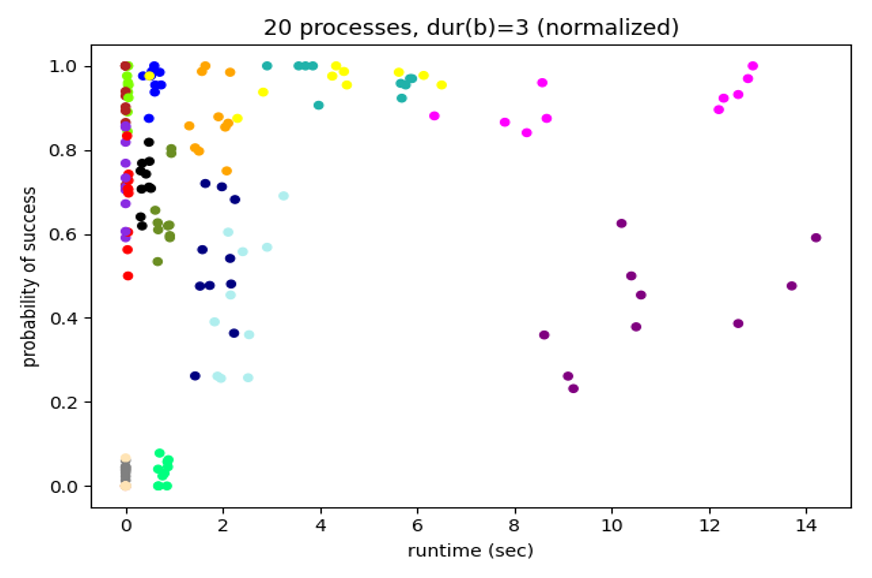}
\includegraphics[width=0.45\columnwidth]{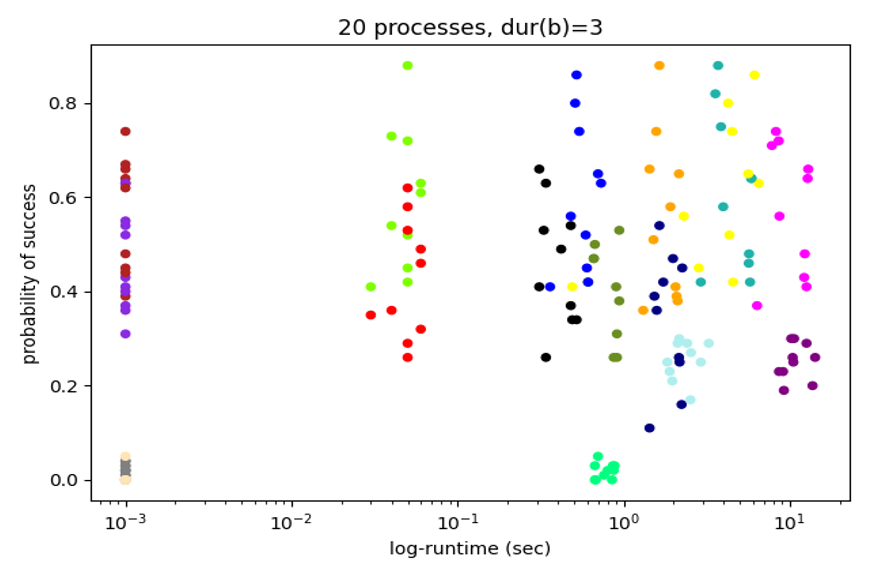}
\includegraphics[width=0.45\columnwidth]{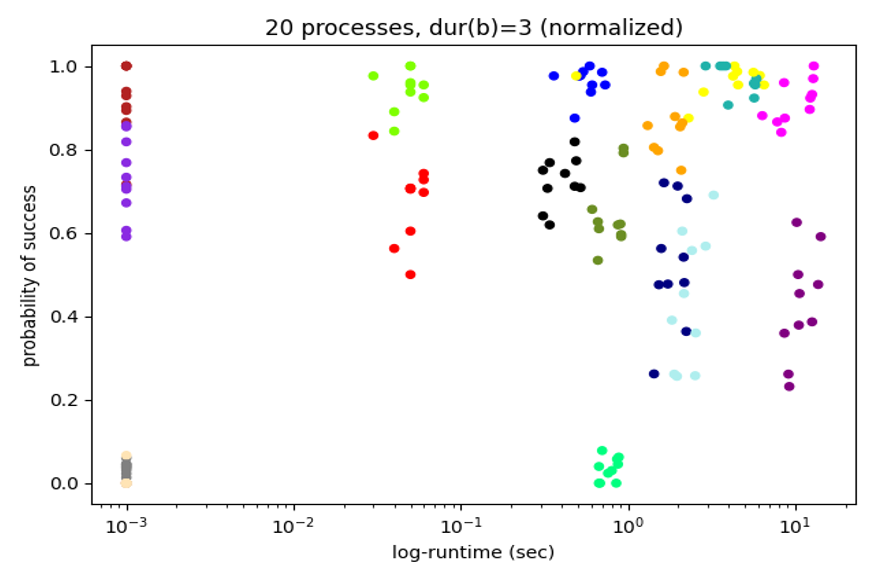}
\end{center}
\caption{All figures present all the results (without averaging). The x-axis stands for the runtime and the y-axis stands for the empirical probability of success. The upper plots present the runtime in linear scale, and the lower plots in logarithmic scale. On the left the results are shown in their raw form, while on the right they are normalized (by the maximal values).}
\label{N=20:all}
\end{figure}

\begin{figure}[ht]
\begin{center}
\includegraphics[width=0.45\columnwidth]{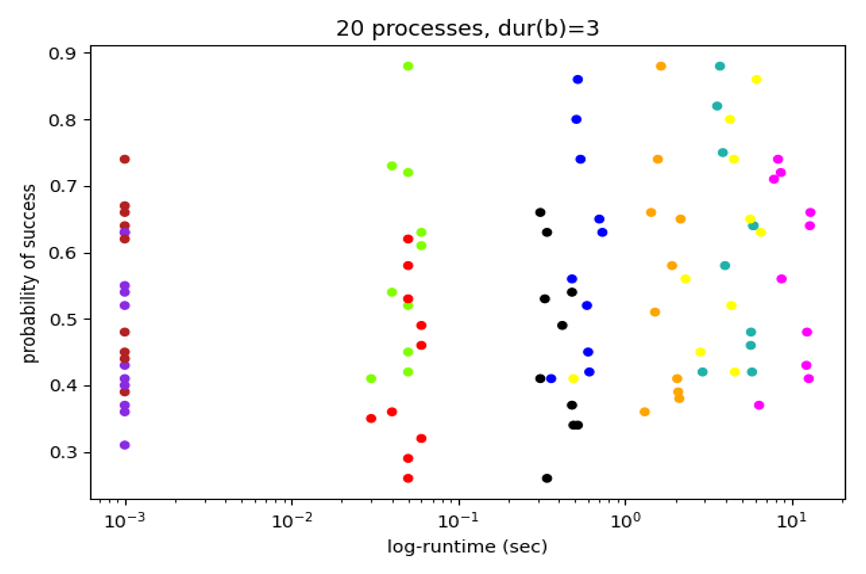}
\includegraphics[width=0.45\columnwidth]{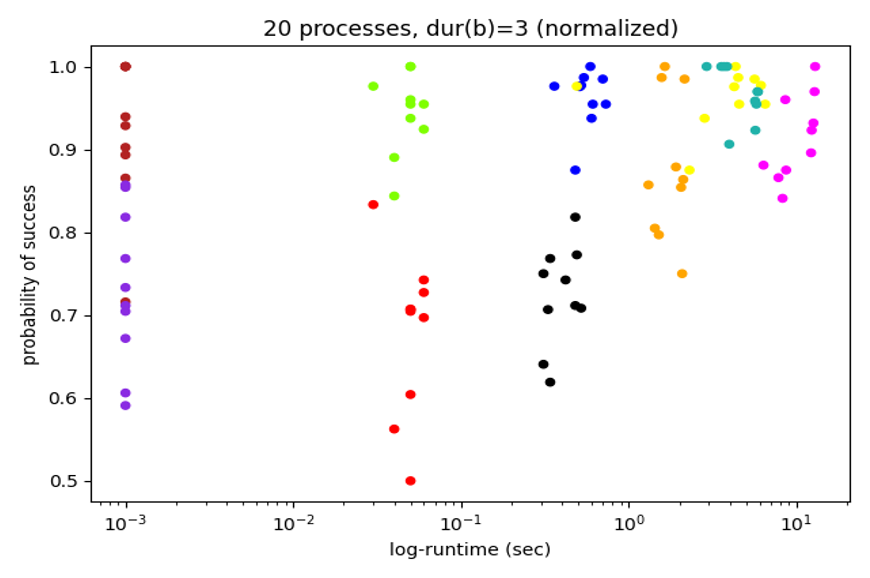}
\end{center}
\caption{The figures present only the 10 algorithms with best probability of success. In both figures the runtime is in log-scale. On the right the results are normalized, and on the left they are not.}
\label{N=20:top10}
\end{figure}

\begin{figure}[ht]
\begin{center}
\includegraphics[width=.45\columnwidth]{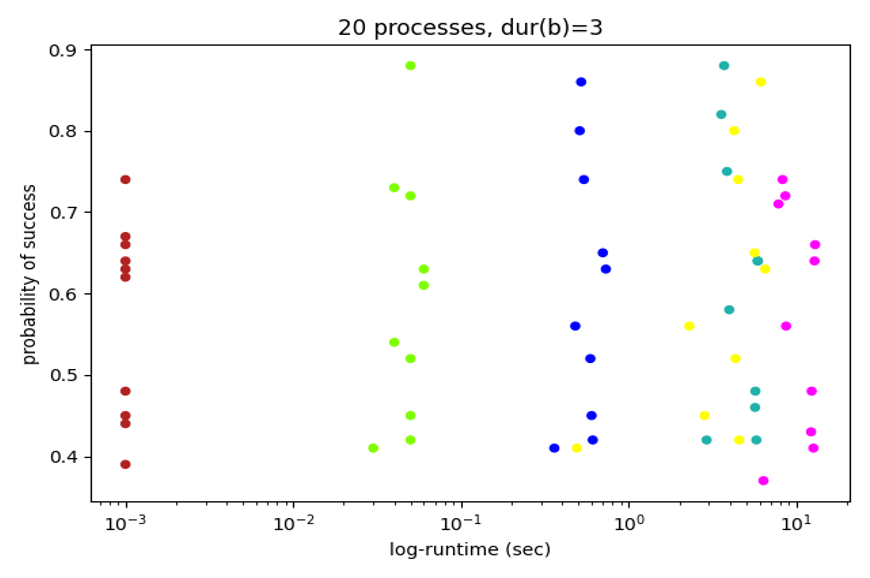}
\includegraphics[width=.45\columnwidth]{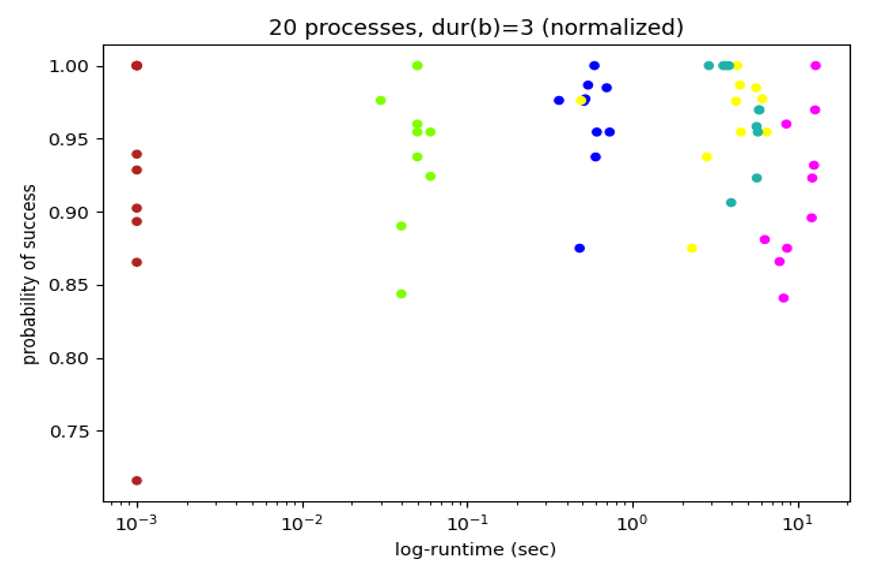}
\end{center}
\caption{The figures present only the 6 algorithms with the best probability of success.}
\label{N=20:top6}
\end{figure}

\begin{figure}[ht]
\begin{center}
\includegraphics[width=0.4\columnwidth]{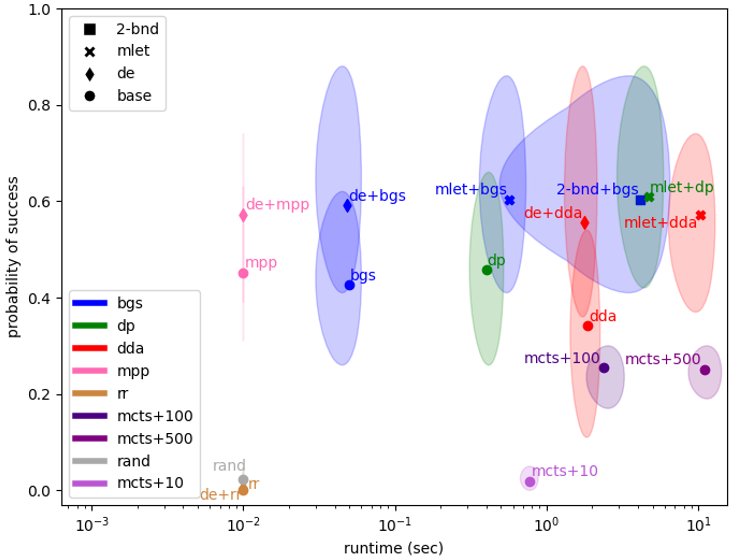}
\end{center}
\caption{Average results for instances with 20 processes and $dur(b)=3$. Ellipse centers are averages over instances, the shaded area just covers the results for all 10 instances.}
\label{N=20:dur=3:ellipses}
\end{figure}

\begin{figure}[ht]
\begin{center}
\includegraphics[width=0.4\columnwidth]{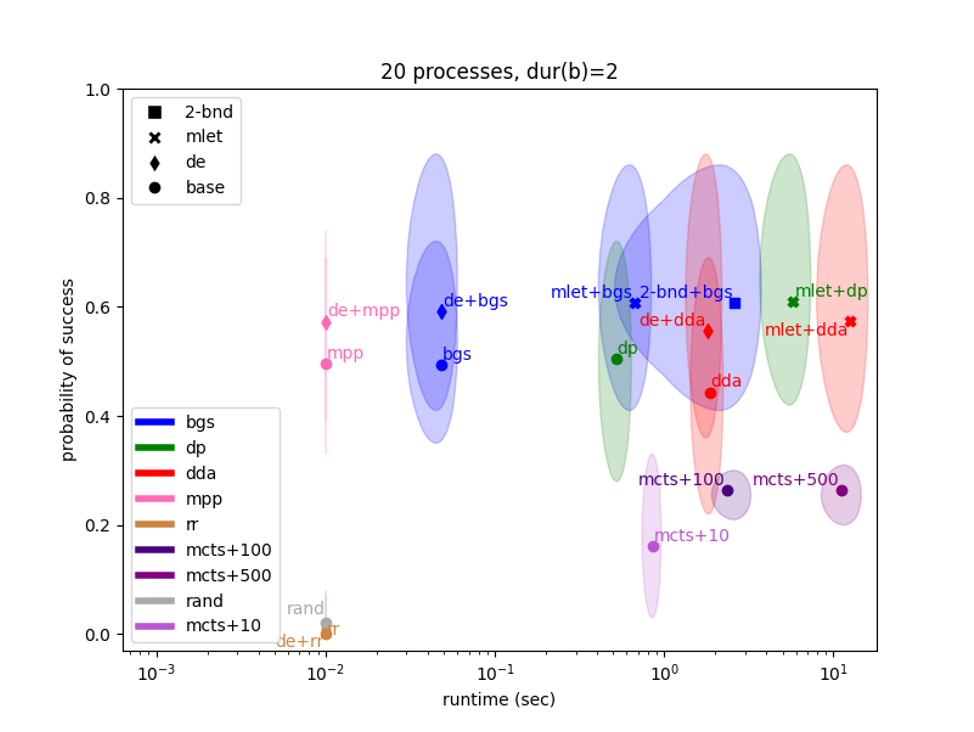}
\end{center}
\caption{Average results for instances with 20 processes and $dur(b)=2$. Ellipse centers are averages over instances, the shaded area just covers the results for all 10 instances.}
\label{N=20:dur=2:ellipses}
\end{figure}

\begin{figure}[ht]
\begin{center}
\includegraphics[width=0.4\columnwidth]{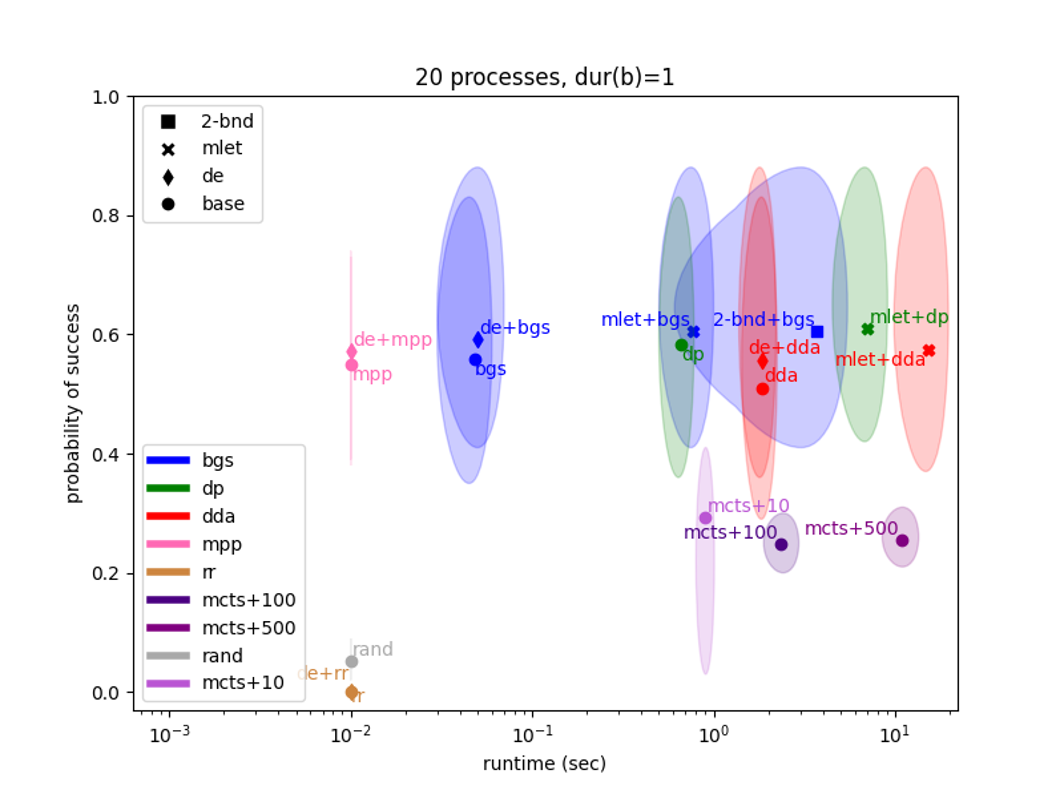}
\end{center}
\caption{Average results for instances with 20 processes and $dur(b)=1$. Ellipse centers are averages over instances, the shaded area just covers the results for all 10 instances.}
\label{N=20:dur=1:ellipses}
\end{figure}

\begin{figure}[ht]
\begin{center}
\includegraphics[width=0.45\columnwidth]{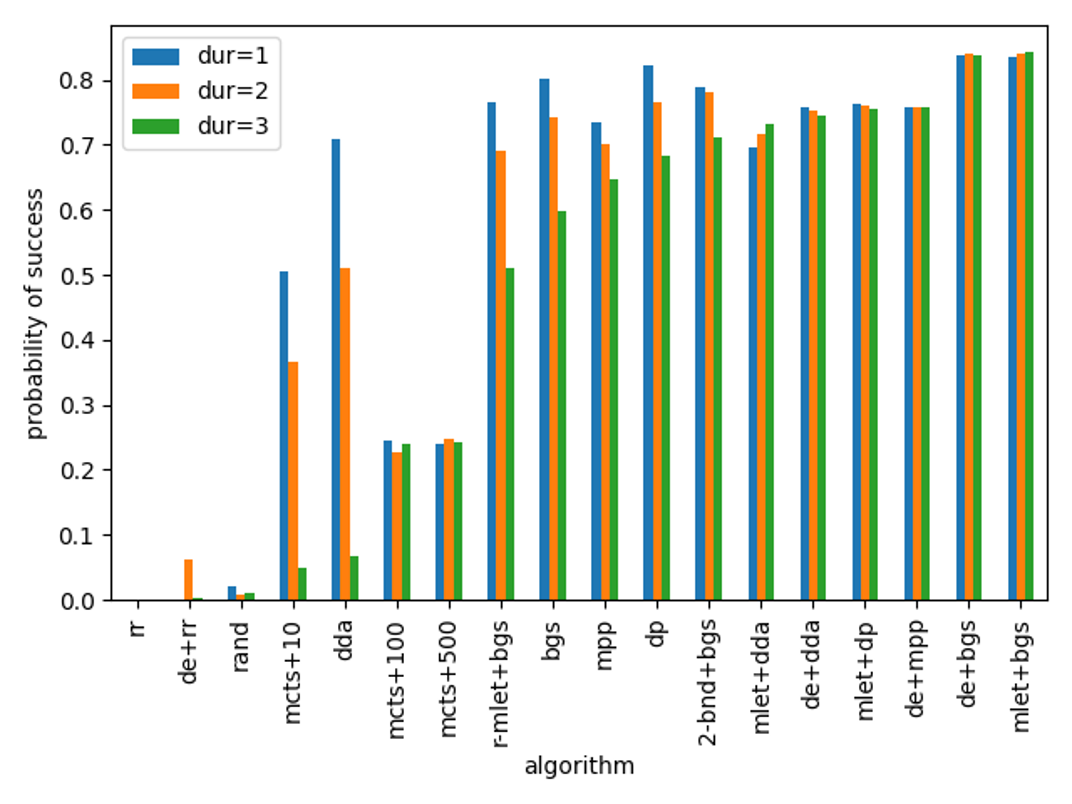}
\includegraphics[width=0.45\columnwidth]{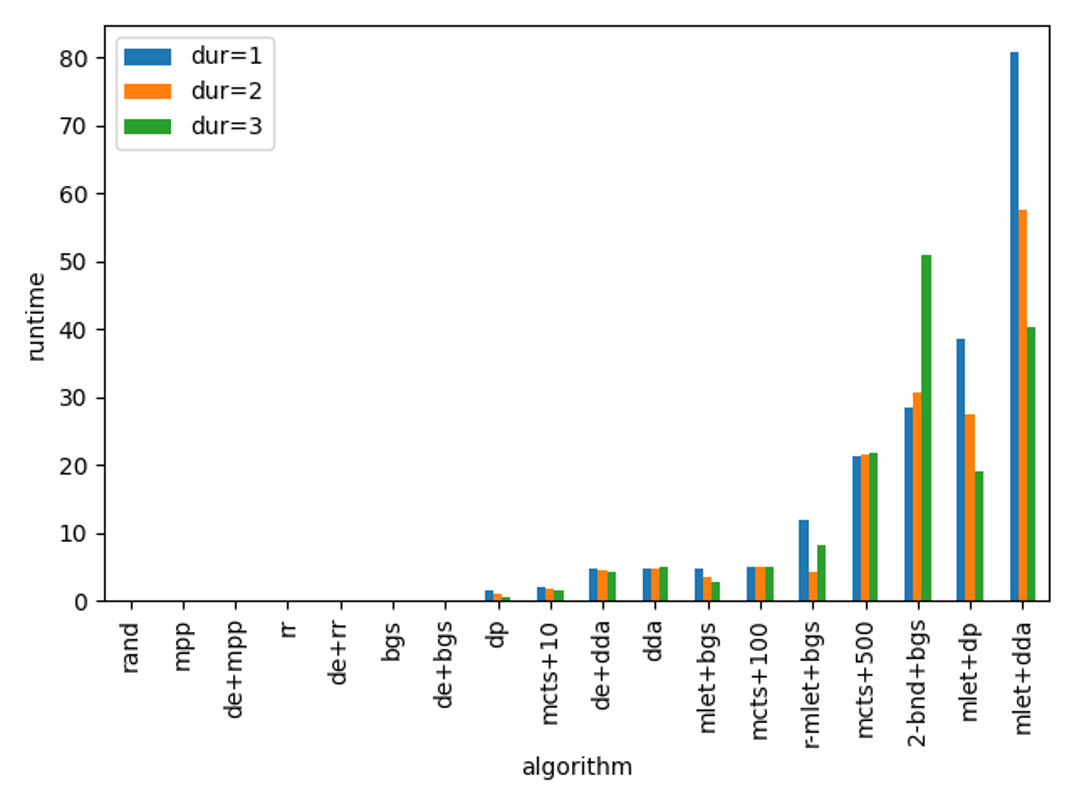}
\end{center}
\caption{Bar plots for 50 processes.}
\label{N=50:bars}
\end{figure}

\begin{figure}[ht]
\begin{center}
\includegraphics[width=0.4\columnwidth]{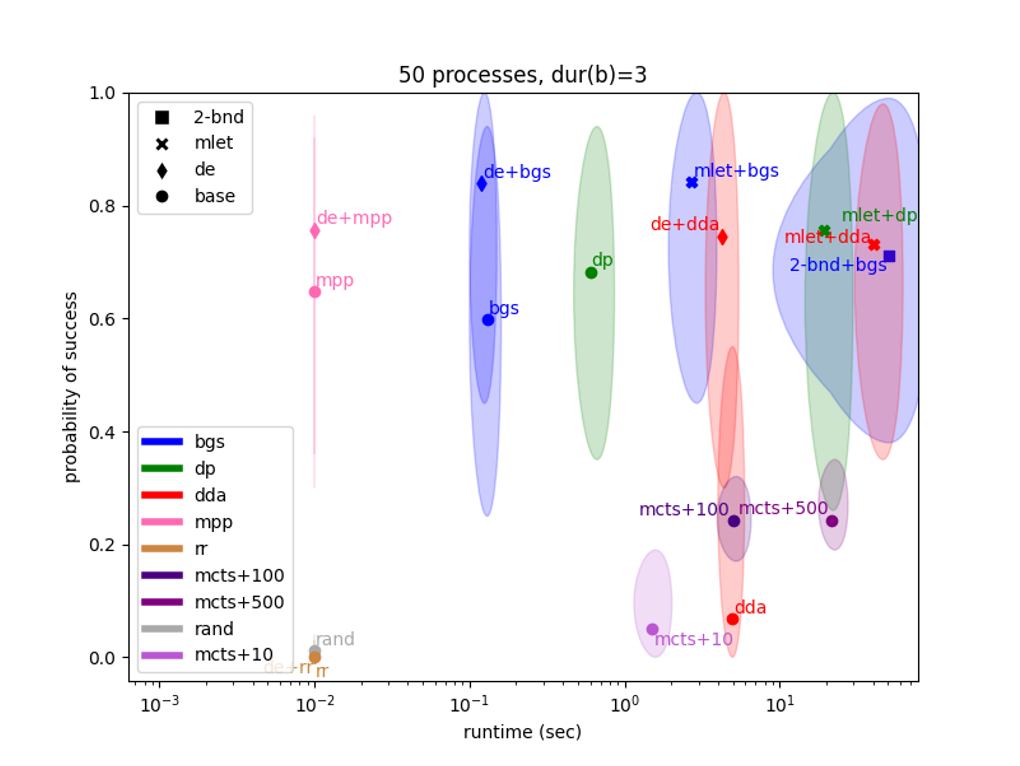}
\end{center}
\caption{Average results for instances with 50 processes and $dur(b)=3$. Ellipse centers are averages over instances, the shaded area just covers the results for all 10 instances.}
\label{N=50:dur=3:ellipses}
\end{figure}

\begin{figure}[ht]
\begin{center}
\includegraphics[width=0.4\columnwidth]{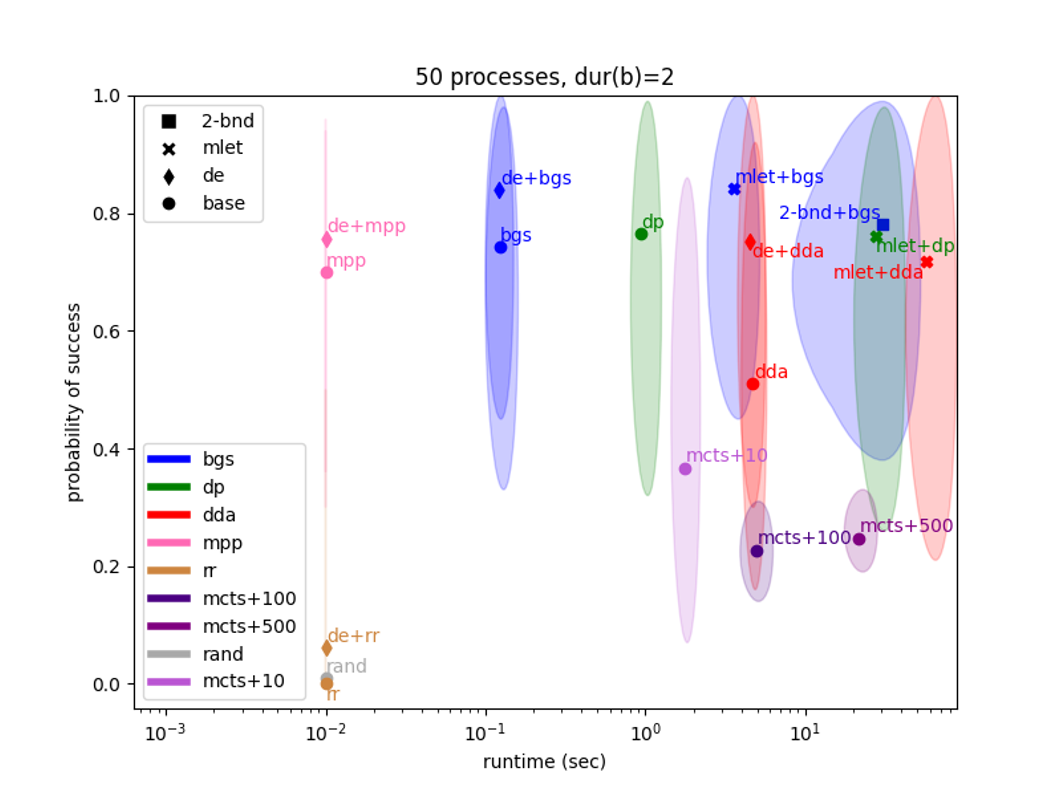}
\end{center}
\caption{Average results for instances with 50 processes and $dur(b)=2$. Ellipse centers are averages over instances, the shaded area just covers the results for all 10 instances.}
\label{N=50:dur=2:ellipses}
\end{figure}

\begin{figure}[ht]
\begin{center}
\includegraphics[width=0.4\columnwidth]{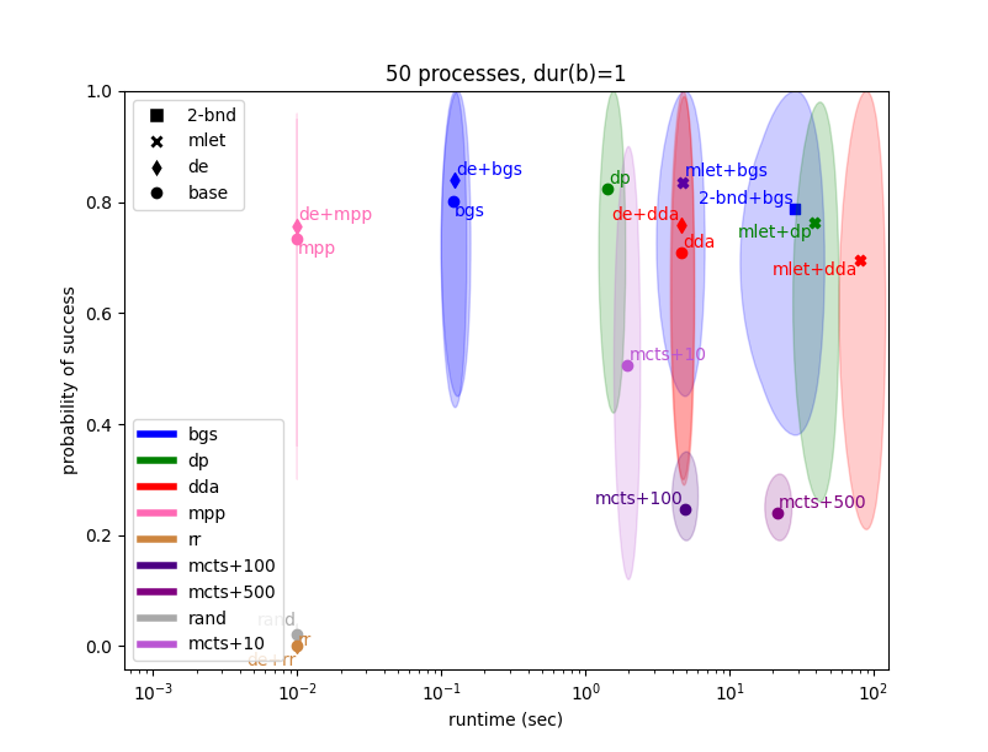}
\end{center}
\caption{Average results for instances with 50 processes and $dur(b)=1$. Ellipse centers are averages over instances, the shaded area just covers the results for all 10 instances.}
\label{N=50:dur=1:ellipses}
\end{figure}

\end{document}